\documentclass[journal,twoside,web]{ieeecolor}

\usepackage{tmi}
\usepackage{cite}
\usepackage{amsmath,amssymb,amsfonts}
\usepackage{algorithmic}
\usepackage{graphicx}
\usepackage{textcomp}
\usepackage{bm}
\usepackage{stackengine,scalerel}
\usepackage{pdfpages}
\usepackage{caption}
\usepackage{subfiles} 
\usepackage{gensymb}
\usepackage{latexsym}
\usepackage{pifont}%
\newsavebox{\bigimage}
\usepackage{framed}
\usepackage{tabulary}
\usepackage{graphicx}
\usepackage{caption}
\usepackage{multicol}
\usepackage{subcaption}
\usepackage{hyperref}

\usepackage{svg}
\usepackage{booktabs}

\usepackage[utf8]{inputenc}
\usepackage[export]{adjustbox}
\usepackage{color}
\usepackage{array}

\newcolumntype{P}[1]{>{\centering\arraybackslash}p{#1}}
\newcolumntype{M}[1]{>{\centering\arraybackslash}m{#1}}

 \definecolor{red}{rgb}{0.0, 0.0, 0.0}
\definecolor{blue}{rgb}{0.0, 0.0, 0.0}
\definecolor{yellow}{rgb}{1.0,1.0,1.0}

\newcommand\overstar[1]{\ThisStyle{\ensurestackMath{%
  \setbox0=\hbox{$\SavedStyle#1$}%
  \stackengine{0pt}{\copy0}{\kern.2\ht0\smash{\SavedStyle*}}{O}{c}{F}{T}{S}}}}

\def\BibTeX{{\rm B\kern-.05em{\sc i\kern-.025em b}\kern-.08em
    T\kern-.1667em\lower.7ex\hbox{E}\kern-.125emX}}
\begin{document}

\title{Uncertainty Estimation for Heatmap-based Landmark Localization}
\author{Lawrence Sch\"obs, Andrew J. Swift, and Haiping Lu, \IEEEmembership{Senior Member, IEEE}
\thanks{This work was supported by EPSRC (2274702) and the Wellcome Trust (215799/Z/19/Z and 205188/Z/16/Z)(Corresponding author: Haiping Lu). }
\thanks{Lawrence Sch\"obs and Haiping Lu are with the Department of Computer Science at the University of Sheffield, Sheffield, S1 4DP, UK (e-mail: laschobs1@sheffield.ac.uk, h.lu@sheffield.ac.uk).}
\thanks{Andrew J. Swift is with the Department of Infection, Immunity \& Cardiovascular Disease, University of Sheffield, UK (e-mail: a.j.swift@sheffield.ac.uk).}}

\maketitle

\begin{abstract}
Automatic anatomical landmark localization has made great strides by leveraging deep learning methods in recent years. The ability to quantify the uncertainty of these predictions is a vital component needed for these methods to be adopted in clinical settings, where it is imperative that erroneous predictions are caught and corrected. We propose Quantile Binning, a data-driven method to categorize predictions by uncertainty with estimated error bounds. Our framework can be applied to any continuous uncertainty measure, allowing straightforward identification of the best subset of predictions with accompanying estimated error bounds. We facilitate easy comparison between uncertainty measures by constructing two evaluation metrics derived from Quantile Binning. {\color{blue} We compare and contrast three epistemic uncertainty measures (two baselines, and a proposed method combining aspects of the two), derived from two heatmap-based landmark localization model paradigms (U-Net and patch-based). We show results across three datasets, including a publicly available Cephalometric dataset. We illustrate how filtering out gross mispredictions caught in our Quantile Bins significantly improves the proportion of predictions under an acceptable error threshold. Finally, we demonstrate that Quantile Binning remains effective on landmarks with high aleatoric uncertainty caused by inherent landmark ambiguity, and offer recommendations on which uncertainty measure to use and how to use it. The code and data are available at \url{https://github.com/schobs/qbin}}.
\end{abstract}


\vspace{-3mm}

\section{Introduction}
\label{sec:introduction}
\IEEEPARstart{A}{utomatic} landmark localization is an important step in many medical image analysis methods, such as image segmentation \cite{1501921} and image registration \cite{johnson2002consistent, murphy2011semi}. An erroneous landmark prediction at an early stage of analysis will flow downstream and compromise the validity of final conclusions. Therefore, the ability to quantify the uncertainty of a prediction is a vital requirement in a clinical setting where explainability is crucial and there is a human in-the-loop to correct highly uncertain predictions \cite{holzinger2016interactive}.

In this study we propose Quantile Binning, a data-driven framework to estimate a prediction's quality by learning the relationship between any continuous uncertainty measure and localization error. Using the framework, we place predictions into bins of increasing subject-level, epistemic uncertainty and assign each bin a pair of estimated localization error bounds. These bins can be used to identify the subsets of predictions with expected high or low localization errors, allowing the user to make a choice of which subset of predictions to review and reannotate based on their expected error bounds. Our method is agnostic to the particular uncertainty metric used, as long as it is continuous and the true function between the uncertainty metric and localization error is monotonically increasing. We showcase our method using three uncertainty measures: a baseline derived from predicted heatmap activations,  {\color{blue} a strong baseline} of ensemble model prediction variance \cite{drevicky2020evaluating}, as well as introducing our own measure based on ensemble heatmap activations. Furthermore, we introduce two uncertainty evaluation methods, measuring how well an uncertainty measure truly predicts localization error and the accuracy of our predicted error bounds.

We explore the efficacy of our three uncertainty metrics on two paradigms of localization models: an encoder-decoder U-Net that regresses Gaussian heatmaps \cite{ronneberger2015u}, and a patch-based network that generates a heatmap from patch voting, PHD-Net \cite{schobs2021confidence}. We compare how the same uncertainty measures perform under the two different approaches to landmark localization on two Cardiac Magnetic Resonance Imaging (CMR) datasets of varying difficulty and a publicly available Cephalometric dataset \cite{wang2016benchmark}, finding promising results for both paradigms. {\color{red} Furthermore, we explore the effect of aleatoric uncertainty caused by landmark ambiguity on Quantile Binning and our uncertainty measures.} Our Quantile Binning method is generalizable to any continuous uncertainty measure, and the examples we investigate in this study can be applied as a post-processing step to any heatmap-based landmark localization method. We aspire that this work can be used as a framework to build, evaluate and compare uncertainty metrics in landmark localization beyond those demonstrated in this paper.
\vspace{-2mm}

\section{Related Work}

\subsection{Landmark Localization}

The recent advancement in machine learning has led to convolutional neural networks (CNNs) dominating the task of landmark localization. Encoder-decoder methods, originally proposed for the task of image segmentation \cite{ronneberger2015u}, have cemented themselves as one of the leading approaches for landmark localization in both the medical domain \cite{payer2019integrating, zhong2019attention, torosdagli2018deep, thaler2021modeling} and computer vision \cite{tang2018quantized, yang2017stacked}. The architecture of these methods allow the analysis of images at multiple resolutions, learning to predict a Gaussian heatmap centred around the predicted landmark location. The activation of each pixel in the heatmap can be interpreted as the pseudo-probability of the pixel being the target landmark. The network learns to generate a high response near the landmark, smoothly attenuating the responses in a small, predefined radius around it. Regressing heatmaps proves more effective than regressing coordinates \cite{zhang2017detecting}, as the heatmap image offers smoother supervision than direct coordinate values, and also models some uncertainty in the prediction.

\begin{figure*}
\includegraphics[]{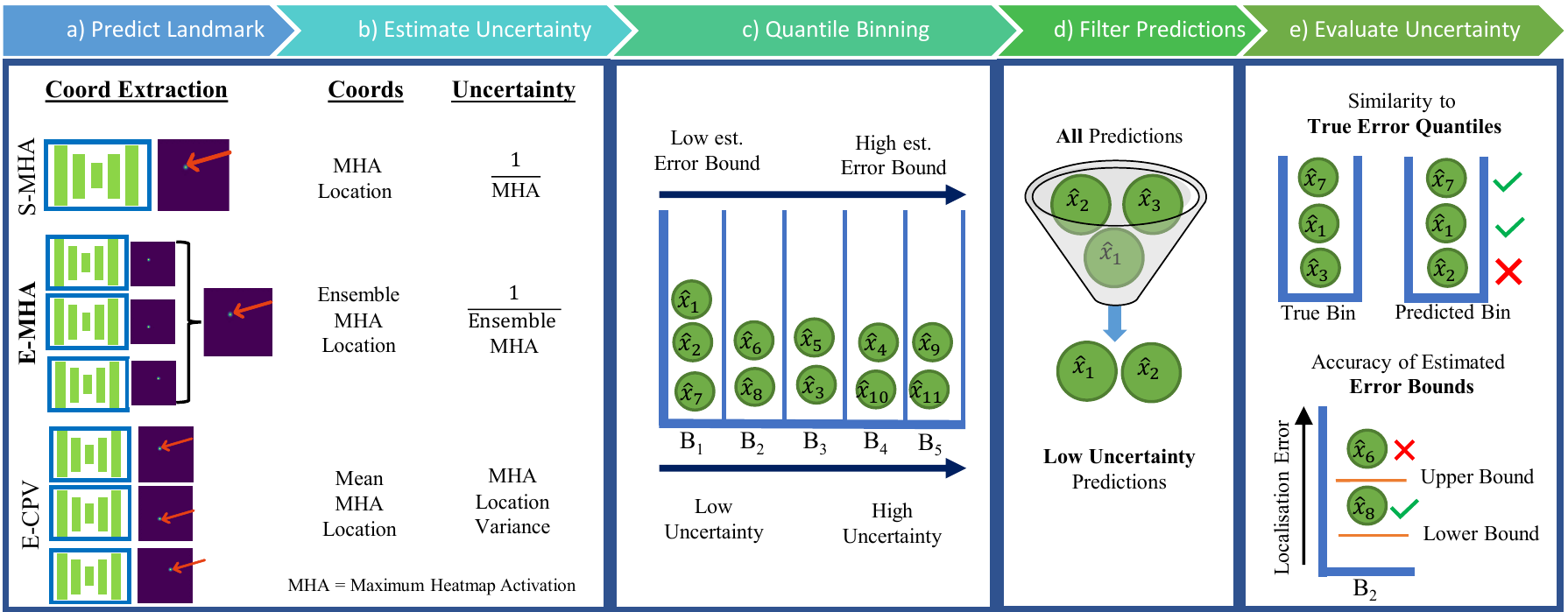}

\caption{Overview of our general Quantile Binning framework. \textbf{a)} We make a prediction using a heatmap-based landmark localization model, and \textbf{b)} extract a continuous uncertainty measure. \textbf{c)} We learn thresholds to categorize predictions into bins of increasing uncertainty, estimating error bounds for each bin. \textbf{d)} We filter out predictions from high uncertainty bins to improve the proportion of acceptable predictions. \textbf{e)} Finally, we evaluate each uncertainty measure's ability to capture the true error quantiles and the accuracy of the estimated error bounds.  }
\vspace{-5mm}

\label{figure:overview}
\end{figure*}
However, in medical imaging the number of available training samples is often small so the encoder-decoder network is forced to be shallow, compromising its performance \cite{zhang2017detecting}. One method to overcome this is via a \textit{patch-based} approach; alleviating the problem by sampling many small ``patches'' from an image, learning the relationship between each patch and the target landmark \cite{emad2015automatic,li2018fast}. This approach can generate orders of magnitude more training samples from a single image compared to the encoder-decoder style methods. Furthermore, patch-based models that use  Fully Convolutional Networks (FCN) have fewer parameters than encoder-decoder architectures, decreasing computational requirements and training times \cite{schobs2021confidence}.

Noothout \textit{et al.} \cite{noothout2018cnn} implemented a patch-based network using an FCN to jointly perform classification and regression on each patch. The coarse binary classification task determines whether a patch contains the landmark, and the more precise regression task estimates the displacement from the patch to the landmark. This multi-task, joint learning leads to a light-weight network and enhanced localization performance, with the two tasks sharing a feature representation that improves the performance of both \cite{zhang2021survey}. However, the resulting network has a strong local focus and is also susceptible to failure if the predicted patch from the classification task is incorrect. In a follow-up work, Noothout \textit{et al.} \cite{noothoutnew} extended their work \cite{noothout2018cnn} into a two-stage method: they first train a CNN to provide global estimates for the landmarks, then employ specialised CNNs for each landmark for the final prediction. This method improves upon the first in terms of localization error, but has the drawback of requiring multiple training stages.

To mitigate the inherent local focus of the patch-based methods, we extended the patch-based network \cite{noothout2018cnn} by borrowing heatmap regression from the encoder-decoder networks; reforming the binary classification task as a Gaussian heatmap regression task \cite{schobs2021confidence}. Named PHD-Net (Patch Heatmap \& Displacement regression), this smoother supervision improved performance, reducing misidentifications compared to using the classification branch from the prior work \cite{noothout2018cnn}. Furthermore, we introduced the method Candidate Smoothing, combining the features from the two branches to output more accurate predictions along with an uncertainty measure.

\vspace{-3mm}
\subsection{Uncertainty Estimation}

Estimating the uncertainty of machine learning predictions is a topic of growing interest, particularly relevant in the domain of medical imaging where there is often a  human in the loop to manually correct flagged predictions. A concentrated effort in uncertainty estimation has been applied to image segmentation by the community, a task similar to landmark localization that instead aims to predict a mask for an entire structure rather than a single point.

{\color{red} Segmentation aims to produce a binary map, with pixels activated on the structure of interest and inactive elsewhere. In traditional heatmap-based landmark localization, only the landmark coordinate has a magnitude of 1, smoothly attenuating to 0 in a set radius. The loss function used by landmark localization is the Mean Squared Error (MSE) between the target and predicted heatmap, whereas segmentation uses pixel-wise classification-based losses \cite{jungo2020analyzing}. Nevertheless, the tasks are similar in that the magnitude of the activation of any given pixel in each image can be leveraged for information on the epistemic ``confidence" (inverse of uncertainty) of the model. Jungo \textit{et al.} \cite{jungo2020analyzing} use pixel activation to measure the uncertainty of each pixel segmentation class, including using the average activation of an ensemble. They found that this naive approach was surprisingly well calibrated and that ensembling $5+$ identical but randomly initialized models significantly improved calibration.} {\color{blue} These results have been corroborated by Mehrtash \textit{et al.} who also used pixel activation for confidence as well as an ensemble of 5 identical, randomly initialized networks to convincing effect \cite{mehrtash2020confidence}.}

{\color{blue} Other successful approaches for epistemic uncertainty estimation use Bayesian Neural Networks \cite{kwon2020uncertainty} or Bayesian approximation methods like Monte-Carlo dropout \cite{nair2020exploring}. However, the prevailing approach is an ensemble of identical, randomly initialized networks.} This method affords better performance \cite{mehrtash2020confidence} and a more accurate mechanism for Bayesian marginalization \cite{wilson2020bayesian} compared to a single model using Monte-Carlo dropout. {\color{red} Fort \textit{et al.}'s study suggests this behaviour is due to random initializations exploring entirely different modes of the loss landscape, facilitating a powerful decorrelation effect between models \cite{fort2019deep}. We see extensive use of ensembles of identical models with random initializations in the domain of medical image segmentation to improve accuracy and estimate uncertainty \cite{jungo2020analyzing, mehta2021qu, mehrtash2020confidence, KARIMI2019186}.  }

In landmark localization we are ultimately predicting a single coordinate point rather than a mask, but similar uncertainty estimation approaches can be utilized. However, there are limited works exploring uncertainty in landmark localization. Payer \textit{et al.} \cite{payer2019integrating} directly modeled aleatoric uncertainty during training by learning the isotropic Gaussian covariances of target heatmaps, and predicting the distribution of likely locations of the landmark at test time. {\color{red} Thaler \textit{et al.} took this approach further, learning anisotropic (directionally skewed) Gaussian heatmaps for each landmark, demonstrating that the learned heatmap shapes correspond to inter-observer variability from multiple annotators \cite{thaler2021modeling}. }In terms of epistemic uncertainty, Lee \textit{et al.} \cite{lee2020automated} borrowed from image segmentation approaches by proposing a Bayesian CNN that utilized Monte-Carlo dropout to predict the location and subject-level uncertainty of cephalometric landmarks.

Another method to measure the subject-level, epistemic uncertainty of a heatmap-based landmark prediction is to measure the maximum heatmap activation (MHA) of the predicted heatmap. Since the activation of a Gaussian heatmap at a particular pixel represents the pseudo-probability of the pixel being the landmark, we can use this pseudo-probability as an uncertainty measure: the higher the activation, the more certain the prediction. Drevicky \textit{et al.} \cite{drevicky2020evaluating} compared MHA with ensemble and Monte-Carlo dropout methods, finding MHA surprisingly effective given its simplicity. However, similarly to image segmentation, they found using an ensemble of models was best at predicting uncertainty. They calculated the coordinate prediction variance between an ensemble of models, and found this method performed best at estimating prediction uncertainty. 

In our earlier work utilising the patch-based model PHD-Net, MHA was also used as the uncertainty metric \cite{schobs2021confidence}. However, the heatmap analysed is distinctly different from the heatmaps predicted by encoder-decoder networks. Rather than explicitly learning a Gaussian function centred around the landmark, the approach combined patch-wise heatmap and displacement predictions. We produced a new non-Gaussian heatmap, where the activation of each pixel is defined by the number of patches that voted for it, regularized by the coarse global likelihood prediction. Therefore, the resulting heatmap represents patch-wise ensemble votes rather than a Gaussian function, where the MHA is the pixel with the most ``patch votes''.

To the best of our knowledge, no study has investigated how heatmap-based uncertainty estimation measures can be used to filter out poor predictions in landmark localization. Furthermore, no general framework has been proposed to compare how well uncertainty measures can predict localization error - an important practical application in clinical settings.

\vspace{-1mm}
\section{Contributions}
\vspace{-1mm}
In this paper, we propose a general framework to compare and evaluate uncertainty measures for landmark localization. This work extends the analysis of MHA in \cite{schobs2021confidence}, with more in depth experiments and comparisons. Our contributions, depicted in Fig. \ref{figure:overview}, are threefold: 

\begin{itemize}

    \item We propose Quantile Binning, a method to categorize predictions by any continuous uncertainty measure, and estimate error bounds for each bin (Fig. \ref{figure:overview}c, Sec. \ref{sec:quantile_binning}).
    
    \item We construct two evaluation metrics for uncertainty estimation methods from Quantile Binning: 1) Similarity between predicted bins and true error quantiles; 2) Accuracy of estimated error bounds (Fig. \ref{figure:overview}e, Sec. \ref{sec:evaluation_methods}).
    
      \item {\color{blue} We evaluate three heatmap-derived uncertainty measures and recommend our proposed method Ensemble Maximum Heatmap Activation (E-MHA) to extract landmark coordinates from an ensemble of heatmaps and estimate uncertainty (Fig. \ref{figure:overview}a,  \ref{figure:overview}b, Sec. \ref{sec:uncertainty_measures}).}

\end{itemize}

We demonstrate the impact of our contributions by using our proposed Quantile Binning to compare E-MHA to two existing coordinate extraction and uncertainty estimation methods: a weak baseline of Single Maximum Heatmap Activation (S-MHA), and a stronger baseline of Ensemble Coordinate Prediction Variance (E-CPV). In Sec.\ref{section:baseline_results}, we compare the baseline coordinate extraction performance of the three approaches, followed by the uncertainty estimation performance in Sec. \ref{section:quantile_bins_analysis}. We explore the reach of heatmap-based uncertainty measures by demonstrating they are applicable to both U-Net regressed Gaussian heatmaps and patch-based voting heatmaps. We show each uncertainty measure can identify a subset of predictions with significantly lower mean error than the full set by filtering out predictions from high uncertainty bins (Fig. \ref{figure:overview}c). {\color{red} In Sec. \ref{section:generalizability} we demonstrate the generalizability of our method by applying Quantile Binning to a publicly available Cephalometric dataset \cite{wang2016benchmark}, with significantly more annotated landmarks and images containing some repetitive structures. We show the flexibility of our method by reporting results over a range of binning resolutions in Sec. \ref{section:comparing_q_values}. Furthermore, in Sec. \ref{section:aleatoric_uncertainty} we select a subset of landmarks from the Cephalometric dataset with multiple annotations (provided by \cite{thaler2021modeling}) to explore the effect of aleatoric uncertainty caused by landmark ambiguity on Quantile Binning using our three uncertainty measures.} Finally, in Sec. \ref{sec:recommendations} we make recommendations for which uncertainty measure to use, and how to use it. 

{\color{blue} We provide an open source implementation of this work and the tabular data obtained from the landmark localization models to reproduce our results, alongside extensive experimental results at \url{https://github.com/schobs/qbin}}. 

\vspace{-3mm}

\section{Methods}

\subsection{Landmark Localization Models}

First, we briefly review the two models we will use for landmark localization, allowing us to compare the generalizability of our uncertainty measures across different heatmap generation approaches. We implement a variation of the popular encoder-decoder networks that regresses Gaussian heatmaps, U-Net \cite{ronneberger2015u}. We also implement a patch-based method, PHD-Net \cite{schobs2021confidence}, which produces a heatmap from patch votes. 

\subsubsection{Encoder-Decoder Model (U-Net)}
The vast majority of state-of-the-art landmark localization approaches are based on the foundation of a U-Net style encoder-decoder architecture. The architecture of U-Net follows a ``U'' shape, first extracting features at several downsampled resolutions, before rebuilding to the original dimensionality in a symmetrical upsampling path. Skip connections are employed between each level, preserving spatial information. The rationale behind the architecture design is to inject some inductive bias into the model architecture itself, helping it learn the local characteristics of each landmark, while preserving the global context.

Rather than regressing coordinates directly, the objective of the model is to learn a Gaussian heatmap image for each landmark, with the centre of the heatmap on the target landmark. For each landmark $L_{i}$ with 2D coordinate position $\widetilde{\mathbf{x}}_{i}$, the 2D heatmap image is defined as the 2D Gaussian function:
\vspace{-3mm}

\begin{equation} \label{equation:gaussian}
g_{i}\left(\mathbf{x} \mid \mid\boldsymbol{\mu}= \widetilde{\mathbf{x}}_{i} ; \sigma\right)=\frac{1}{(2 \pi)  \sigma^{2}} \exp \left(-\frac{\left\|\mathbf{x}-\boldsymbol{\mu}\right\|_{2}^{2}}{2 \sigma^{2}}\right),
\end{equation}

\noindent where $\mathbf{x}$ is the 2D coordinate vector of each pixel and $\sigma$ is a user-defined standard deviation. The network learns weights $\mathbf{w}$ and biases $\mathbf{b}$ to predict the heatmap $h_{i}(\mathbf{x} ; \mathbf{w}, \mathbf{b})$. During inference, we can interpret the activation of each pixel in the predicted heatmap as the pseudo-probability of that pixel being the landmark. We will exploit this in our uncertainty estimation methods.

\subsubsection{Patch-based Model (PHD-Net)}
Patch-based models use a Fully Convolutional Network (FCN), with the architecture resembling the first half of an encoder-decoder architecture. Therefore, they are more light-weight than encoder-decoder networks, with significantly less parameters leading to faster training.
 
In our earlier work, we proposed PHD-Net: a multi-task patch-based network \cite{schobs2021confidence}, building on the work by Noothout \textit{et al.} \cite{noothout2018cnn}. We incorporate a variant of the heatmap objective function from encoder-decoder networks into the objective function, predicting the 2D  displacement from each patch to the landmark alongside the coarse Gaussian pseudo-probability of each patch.

PHD-Net aggregates the patch-wise predictions to obtain a heatmap by plotting candidate predictions from the displacement branch as small Gaussian blobs, then regularising the map by the upsampled Gaussian from the heatmap branch. \par

Again, we can consider the activation of each pixel in heatmap as an indicator for uncertainty, where instead of the pseudo-probability, the activation represents the amount of ``patch votes''.

\subsubsection{Ensemble Models}
Using an ensemble of identical but randomly initialized models is more robust than using a single model, as it reduces the effect of a single model becoming stuck in a local minima during training.  {\color{red} Furthermore, random initializations explore different modes of the loss landscape, facilitating a powerful decorrelation effect between models \cite{fort2019deep}}. We use the variance in the predictions of each model to estimate the uncertainty of the prediction, using an use an ensemble of $T$ models.

\vspace{-3mm}

\subsection{Estimating Uncertainty and Coordinate Extraction}
\label{sec:uncertainty_measures}
Although generated differently, we hypothesize both U-Net and PHD-Net produce heatmaps containing useful information to quantify a prediction's uncertainty - but are they equally effective? To this end, we compare the performance of both models under three uncertainty estimation methods: two baseline approaches, and a proposed approach extending one of the baselines to an ensemble of networks. Each method extracts coordinate values from the predicted heatmap, and estimates the prediction's uncertainty.

\subsubsection{Single Maximum Heatmap Activation (S-MHA)}
We introduce the baseline coordinate extraction and uncertainty measure. We use the standard method to obtain the predicted landmark's coordinates $\widehat{\mathbf{x}}_{i}$ from the predicted heatmap  $h_{i}(\mathbf{x} ; \mathbf{w}, \mathbf{b})$, by finding the pixel with the highest activation: 
\vspace{-2mm}

\begin{equation}
\widehat{\mathbf{x}}_{i}=\arg \max _{\mathbf{x}} h_{i}(\mathbf{x} ; \mathbf{w}, \mathbf{b}).
\label{equation:argmax}
\end{equation}

We hypothesize that the pixel activation at the coordinates $\widehat{\mathbf{x}}_{i}$ can describe the model's uncertainty: the higher the activation, the lower the uncertainty, and the lower the prediction error. However, due to this inverse relationship, this measures ``confidence'', not uncertainty.

We transform our confidence metric to an ``uncertainty'' metric $\widehat{y}_{i}$, by applying the following transformation to the pixel activation at the predicted landmark location:
\vspace{-1mm}
\begin{equation}
    \widehat{y}_{i} = \frac{1}{ \underset{\mathbf{x}}{\max}\,  h_{i}(\mathbf{x} ; \mathbf{w}, \mathbf{b})+\epsilon},
\label{eq:conftransform}
\vspace{-2mm}
\end{equation}
where $\epsilon$ is a small constant scalar that prevents $ \frac{1}{0}$. Now, as the pixel activation at $\widehat{\mathbf{x}}_{i}$ increases,  $\widehat{y}_{i}$ decreases.

We call the transformed activation of this peak pixel Single Maximum Heatmap Activation (S-MHA). This is a continuous value bounded between $[\frac{1}{\epsilon}, \frac{1}{1+\epsilon}]$ for U-Net, and bounded between $[\frac{1}{\epsilon}, \frac{1}{N+\epsilon}]$ for PHD-Net, where $N$ is the number of patches. The lower the S-MHA, the lower the uncertainty.

\subsubsection{Ensemble Maximum Heatmap Activation (E-MHA)}
In this work we extend the S-MHA uncertainty measure to ensemble models. We hypothesize that E-MHA should hold a stronger correlation with error than S-MHA due to the additional robustness an ensemble of models affords. We generate the mean heatmap of the $T$ models in the ensemble, and obtain the predicted landmark coordinates as the pixel with the highest activation:

\vspace{-4mm}

\begin{equation}
\widehat{\mathbf{x}}_{i}=\arg \max _{\mathbf{x}} \frac{1}{T} \sum_{t=1}^{T}  h_{i}^{t}(\mathbf{x} ; \mathbf{w}, \mathbf{b}).
\label{equation:EMHA}
\end{equation}
\vspace{-3mm}

{\color{red} Using the average prediction of an ensemble is the simplest, low-cost, standard form of ensemble fusion \cite{jungo2020analyzing, mehta2021qu, mehrtash2020confidence, KARIMI2019186}.} Again, we hypothesize the activation of the pixel {\color{blue} $\widehat{\mathbf{x}}_{i}$} correlates with model confidence. Similar to S-MHA, we inverse the pixel activation and add a small $\epsilon$ to the activation of $\widehat{\mathbf{x}}_{i}$ to give us our uncertainty measure, $\widehat{y}_{i}$:
\vspace{-1mm}

\begin{equation}
    \widehat{y}_{i} = \frac{1}{ \left(\underset{\mathbf{x}}{\max}\,  \frac{1}{T} \sum_{t=1}^{T}  h_{i}^{t}(\mathbf{x} ; \mathbf{w}, \mathbf{b})\right)+\epsilon}.
\label{eq:conftransform}
\end{equation}
\vspace{-1mm}

E-MHA is a continuous value constrained to the same bounds as S-MHA. This is a form of late feature fusion, combining features from all models before a decision is made.

\subsubsection{Ensemble Coordinate Prediction Variance (E-CPV)}
We also implement an additional strong baseline for uncertainty estimation: Ensemble Coordinate Prediction Variance (E-CPV) \cite{drevicky2020evaluating}. The more the models disagree on where the landmark is, the higher the uncertainty.

To extract a landmark's coordinates we first use the traditional S-MHA coordinate extraction method on each of the $T$ models' predicted heatmaps. Then, we use decision-level fusion to calculate the mean coordinate of the individual predictions to compute the final coordinate predictions $\widehat{\mathbf{x}}_{i}$:
\vspace{-2mm}

\begin{equation}
\widehat{\mathbf{x}}_{i} =\frac{1}{T} \sum_{t=1}^{T} \arg \max _{\mathbf{x}} h_{i}^{t}(\mathbf{x} ; \mathbf{w}, \mathbf{b}).
\end{equation}
\vspace{-2mm}

We generate the E-CPV by calculating the mean absolute difference between the $T$ model predictions $\left(\widehat{\mathbf{x}}_{i}^{1}  \text { to } \widehat{\mathbf{x}}_{i}^{T}\right)$ and $\widehat{\mathbf{x}}_{i}$:
\vspace{-3mm}
\vspace{-2mm}

\begin{equation}
\widehat{y}_{i} =\frac{1}{T} \sum_{t=1}^{T}\left\|\widehat{\mathbf{x}}_{i}^{t}-\widehat{\mathbf{x}}_{i}\right\|.
\end{equation}
\vspace{-2mm}

This is a continuous value bounded between $[0,\sqrt{H^{2}+W^{2}}]$, where $H$ and $W$ are the height and width of the original image, respectively. The more the models disagree on the landmark location, the higher the coordinate prediction variance, and the higher the uncertainty.

{\color{blue} Unlike S-MHA and E-MHA, this metric completely ignores the value of the heatmap activations. Therefore, it potentially loses useful uncertainty information but avoids possible bias caused by model miscalibration \cite{guo2017calibration} or the Gaussian assumptions of the target heatmap.}

\vspace{-2mm}

\subsection{Quantile Binning: Categorising Predictions by Uncertainty and Estimating Error Bounds}
\label{sec:quantile_binning}

We leverage the described uncertainty measures to inform the epistemic, subject-level uncertainty of any given prediction, i.e. \textit{is the model's prediction likely to be accurate, or inaccurate based on this uncertainty value?} We propose a data-driven approach, Quantile Binning, using a hold-out validation set to establish thresholds delineating varying levels of uncertainty specific to each trained model. We use these learned thresholds to categorize our predictions into bins and estimate error bounds for each bin. {\color{blue} We opt for a data-driven approach over using static, pre-defined thresholds to increase robustness. For example, two identical models with randomly initialized weights trained on the same training set will converge to different modes \cite{fort2019deep}, with a different distribution of MHA on the same test set. Furthermore, the difficulty of the landmark will also influence the characteristics of the resulting localization model as well as the distribution of the uncertainty measures. Therefore, establishing a set of thresholds for each model is more invariant to training differences compared to using the same thresholds for all models.}

Quantile Binning is application agnostic; applicable to any data as long as it consists of continuous tuples of {\color{blue} \small{\texttt{<Uncertainty Measure, Evaluation Metric>}}}. {\color{red} In this context, a continuous tuple is a pair of continuous variables output by the prediction model, relating to a single sample. }

In this paper, we generate these pairings after the landmark localization model is trained. We use a hold-out validation set and make coordinate predictions and uncertainty estimates using each of our three uncertainty measures described in Section \ref{sec:uncertainty_measures}. Since we have the ground truth annotations of the validation set we can produce continuous {\color{blue} \small{\texttt{<Uncertainty Measure, Localization Error>}}} tuples for each uncertainty measure.

\subsubsection{Establishing Quantile Thresholds}

We aim to categorize predictions using our continuous uncertainty metrics into $Q$ bins. We make the following assumption: \textit{The true function between a good uncertainty measure and localization error is monotonically increasing (i.e. the higher the uncertainty, the higher the error).}

Quantile binning is a non-parametric method that fits well with these assumptions - a variant of histogram binning which is commonly used for calibration of predictive models \cite{guo2017calibration, naeini2015obtaining}. By considering the data in quantiles rather than intervals, we can better capture a skewed distribution as the outliers in the tail of the distribution can be grouped into the same group. In other words, \textit{quantiles divide the probability distribution into areas of approximately equal probability.} 
 
This property allows us to interrogate model-specific (epistemic) uncertainties. Rather than compute uncertainty thresholds based on predefined error thresholds for each bin, we use Quantile Binning to create thresholds that group our samples in relative terms. This enables the user to flag the worst $X\%$ of predictions. We describe the steps below.

First, for any given uncertainty measure we sort our validation set {\color{blue} \small{\texttt{<Uncertainty Measure, Localization Error>}}} tuples in ascending order of their uncertainty value and sequentially group them into $Q$ equal-sized bins $B_{1},...,B_{Q}$. {\color{blue} We assign each bin $B_{q}$ a pair of boundaries defined by the uncertainty values of the tuples at the edges of the bin to create an interval: $[\alpha_{q-1}, \alpha_{q})$.} To capture all predictions at the tail ends of the distribution, we set $\alpha_{0} = 0$, and $\alpha_{Q} = \infty$.

During inference, we use these boundaries to bin our predictions into $Q$ bins ($B_{1}$...$B_{Q}$), with uncertainty increasing with each bin. For each predicted landmark $\widehat{\mathbf{x}}_{i}$ with uncertainty $\widehat{y}_{i}$ where $ \alpha_{q-1} \leq \widehat{y}_{i} < \alpha_{q}$,  $\widehat{\mathbf{x}}_{i}$ is binned into  $B_{q}$. {\color{blue} As long as the validation set is representative of the true distribution,} the distribution of samples should be uniform across the bins due to the quantile method we used to obtain thresholds.

The higher the value of $Q$, the more fine-grained we can categorize our uncertainty estimates. However, as $Q$ increases the method becomes more sensitive to any noise present in the uncertainty measure, leading to less accurate prediction binnings. {\color{red} We demonstrate this trade-off in Sec. \ref{section:comparing_q_values}.}

{\color{blue} Since the uncertainty boundaries are defined by the density of the validation set distribution, the method is agnostic to the absolute range of the uncertainty measure. Therefore it is applicable to any continuous uncertainty measure.}

\subsubsection{Estimating Error Bounds using Isotonic Regression}
Establishing thresholds has allowed us to filter predictions by uncertainty in relative terms, but we lack a method to estimate absolute localization error for each bin. For example, for an easy landmark, the samples in $B_{1}$ may have a very low localization error in absolute terms, but for a more difficult landmark even the lowest relative uncertainty samples in $B_{1}$ may have a high error. Therefore, in order to offer users information about the expected error for each group, we present a data-driven approach to predict error bounds. 

{\color{blue} A simple approach would be to observe the localization error of the tuple at the quantile boundaries $[\alpha_{q-1}$ and $\alpha_{q})$. However, observing a single sample from the validation set is subject to noise and may produce a poor estimate for an error bound. Therefore, on our hold-out validation set, we first use Isotonic Regression to approximate the function between uncertainty and error, constraining it to be monotonically increasing.} Isotonic regression is a method to fit a free-form, non-decreasing line to a set of observations, also commonly used for predictive model calibration \cite{zadrozny2002transforming, guo2017calibration}. It is non-parametric, so can learn the true distribution if given enough i.i.d. data. Given a list of $n$ observations  $\left\{\left(\eta_{1}, \beta_{1}\right), \ldots,\left(\eta_{n}, \beta_{n}\right)\right\}$, the regression seeks a weighted least squares fit $\widehat{\beta}_{i}\approx \beta_{i}$ subject to the constraint that $\hat{\beta}_{i} \leq \hat{\beta}_{j} \text { whenever } \eta_{i} \leq \eta_{j}$:
\vspace{-3mm}

\vspace{-3mm}
\begin{equation}
    \min \sum_{i=1}^{n} \left(\hat{\beta}_{i}-\beta_{i}\right)^{2} \text { s.t. } \hat{\beta}_{i} \leq \hat{\beta}_{j} \text { for all }(i, j) \in E,
\end{equation}
\vspace{-2mm}

\noindent where $E = \left\{(i, j): \eta_{i} \leq \eta_{j}\right\}$. In our case, the observations are the $(Uncertainty Measure, \,Localization Error)$ tuples.

Next, we use our isotonically regressed line to estimate error bounds for each of our quantile bins. {\color{blue} We input each bin's threshold intervals [$\alpha_{q-1}$, $\alpha_{q}$) into our fitted Isotonic Regression function, obtaining error predictions for each threshold,  [$\gamma_{q-1}, \gamma_{q})$.} We use these values as the estimated lower and upper error bounds, respectively, of the predictions in bin $B_{q}$. Note, that for $B_{1}$ we only estimate an upper bound, and for $B_{Q}$ we only estimate a lower bound.

In summary, we use a data-driven approach to learn thresholds to progressively filter predictions at inference into $Q$ bins of increasing uncertainty, and assign each bin estimated error bounds.
\vspace{-5mm}

\subsection{Evaluation Metrics for Uncertainty Measures}
\vspace{-1mm}
\label{sec:evaluation_methods}
Next, we construct methods to evaluate how well an uncertainty measure's predicted bins represent the true error quantiles, and how accurate each bin's estimated error bounds are.

\subsubsection{Evaluating the Predicted Bins}

A good uncertainty measure will have a strong correlation with localization error. Therefore, it should provide quantile thresholds that correspond to the true error quantiles. For example, since Bin $B_{1}$ contains the predictions with the uncertainties at the lowest $\frac{1}{Q}$ quantile, the localization errors of the predictions in $B_{1}$ should be the lowest $\frac{1}{Q}$ quantile of the test set. This can be generalized to each group, until $B_{Q}$, which should contain the errors in the $\frac{Q-1}{Q}$ quantile.

To evaluate this desired property, we propose to measure the similarity between each predicted bin and its respective theoretically perfect bin.

We create the ground truth (GT) bins by ordering the test set samples in ascending order of error. Then, we sequentially bin them into $Q$ equally sized bins: $\widehat{B}_{1}...\widehat{B}_{Q}$.

For each predicted and GT bin pair $B_{q}$ \& $\widehat{B}_{q}$, we calculate the Jaccard Index (JI) between them and report the mean measure of each bin across all folds:
\begin{equation}
    J_{q}(B_{q}, \widehat{B}_{q})=\frac{|B_{q} \cap \widehat{B}_{q}|}{|B_{q} \cup \widehat{B}_{q}|}.
\end{equation}
\vspace{-1mm}
The higher the JI, the better the uncertainty measure has binned predictions by localization error. Therefore, it follows that the higher the JI, the better the uncertainty measure predicts localization error.

\subsubsection{Accuracy of Estimated Error bounds}
A good uncertainty measure will have a monotonically increasing relationship with localization error. Therefore, estimating the true function using isotonic regression should provide accurate error bound estimations.

To measure this, for each bin $B_{q}$, we calculate the percentage of predictions whose error falls between the estimated error bound interval, $[\gamma_{q-1}, \gamma_{q})$. The higher the percentage, the higher the accuracy of our estimated error bounds.

\begin{figure}[t!]
    \centering
      \begin{subfigure}[t]{0.32\linewidth}
        \includegraphics[width=\linewidth]{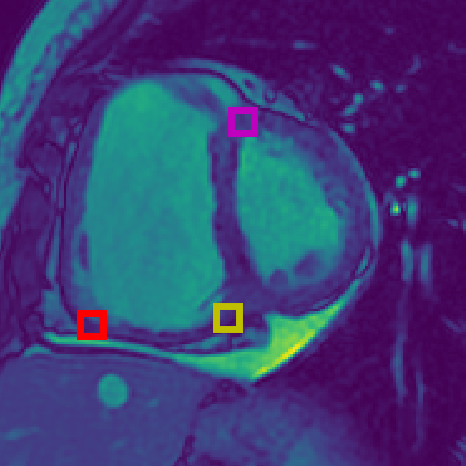}
        \caption{SA CMR.}
        \label{subfig: landmarkSA}
    \end{subfigure}
	\begin{subfigure}[t]{0.32\linewidth} %
        \includegraphics[width=\linewidth]{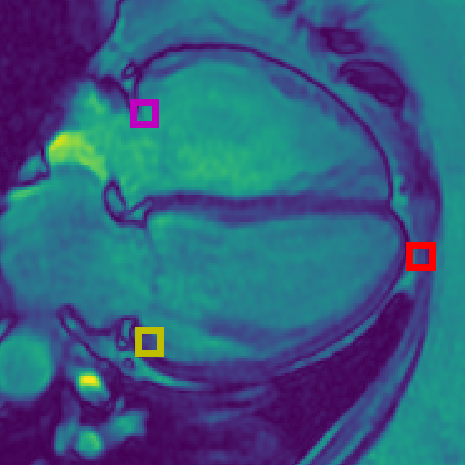}
    	\caption{4CH CMR.}
        \label{subfig:landmark4Ch}
    \end{subfigure}
	\begin{subfigure}[t]{0.32\linewidth} %
        \includegraphics[width=\linewidth]{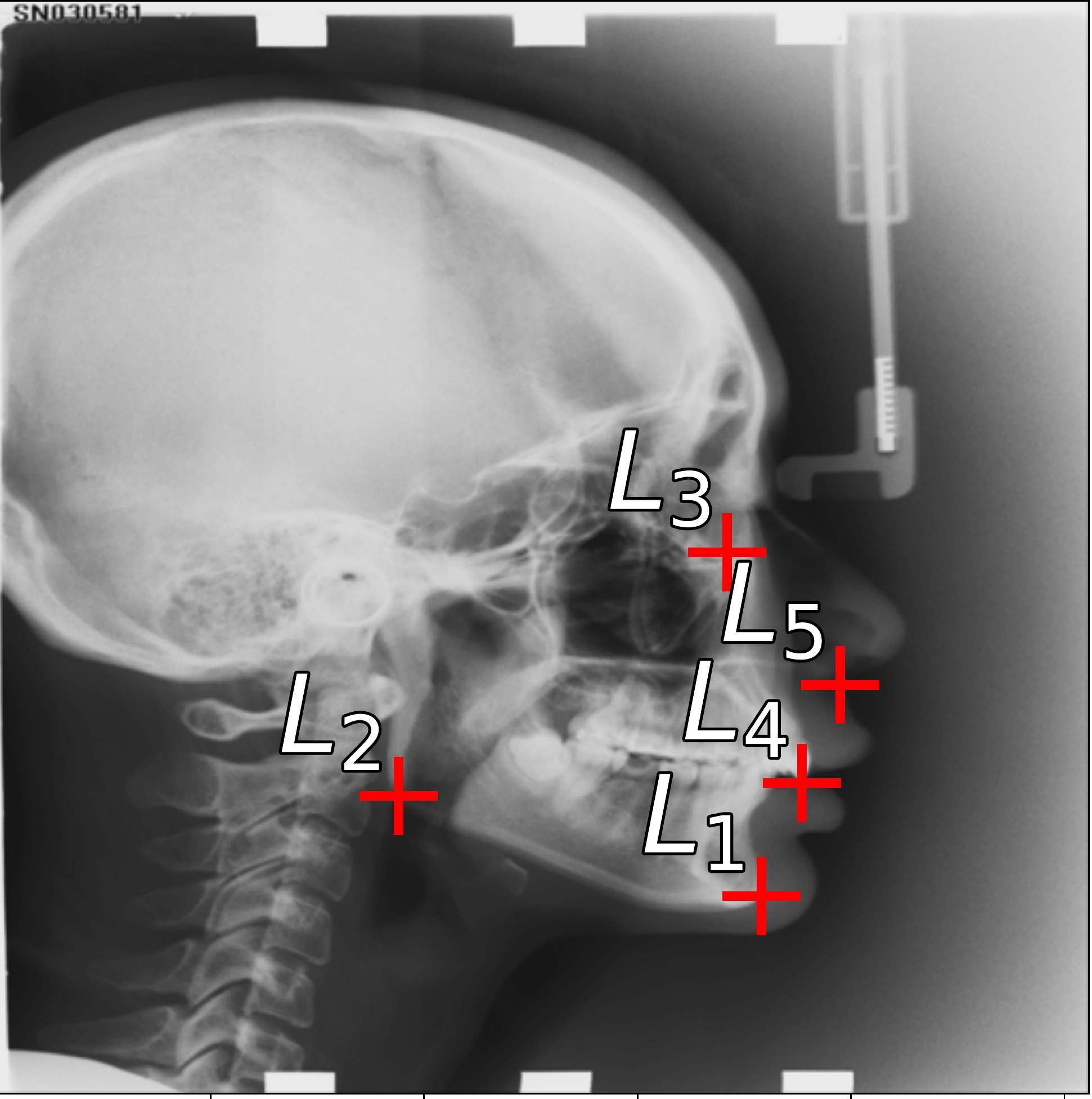}
    	\caption{{\color{red}Cephalometric.}}
        \label{subfig:landmarkisbi}
    \end{subfigure}

\caption{\textbf{(a)} Landmarks for Short Axis (SA) CMR: Magenta = superior right ventricle insertion point valve; Yellow = inferior right ventricle insertion point; Red = inferior lateral reflection of right ventricle free wall. \textbf{(b)} Landmarks for 4 chamber (4CH) CMR: Magenta = tricuspid valve; Yellow = mitral valve; Red = apex of left ventricle. {\color{red}\textbf{(c)} Subset of Landmarks included in the Cephalometric dataset \cite{wang2016benchmark}. Displayed landmarks are used in the aleatoric uncertainty analysis (Sec. \ref{section:aleatoric_uncertainty}).}}
\vspace{-5mm}

\label{fig:cardiac_example}

\end{figure}

\section{Datasets}

{\color{red} We perform our experiments using three datasets. The first two datasets are from the ASPIRE Registry \cite{hurdman2012aspire}, containing Cardiac Magnetic Resonance Imaging (CMR) sequences, from a 1.5 Tesla GE HDx (GE Healthcare, Milwaukee, USA) system using an eight-channel cardiac coil. Images were acquired using a cardiac-gated multislice balanced steady-state free precession sequence (20 frames per cardiac cycle, slice thickness 10mm, 0mm inter-slice gap, field of view 480mm, acquisition matrix $256 \times 200$, flip angle $ 60\degree $, BW 125 KHz/pixel, TR/TE 3.7/1.6 ms).} Each subject has a four chamber (4CH) view and/or a short axis view (SA). Each CMR sequence has a spatial resolution of $512 \times 512$ pixels, where each pixel represents 0.9375mm of the organ, and the first frame was used for landmark localization in this study. There are 303 SA images, each with three annotated landmarks: the inferior right ventricle insertion point (infSA), the superior right ventricle insertion point (supSA), and the inferior lateral reflection of the right ventricle free wall (RVSA). There are 422 4CH images, each with three annotated landmarks: the apex of the left ventricle at end diastole (LVDEV Apex), the mitral valve (mitral), and tricuspid valve (tricuspid). The 4CH dataset represents a more challenging landmark localization task as the images have much higher variability than the SA dataset. The landmarks were decided and manually labelled by a radiologist, as shown in Figs. \ref{subfig: landmarkSA} \& \ref{subfig:landmark4Ch}. For this study, we consider the SA images the \textbf{EASY} dataset, and the 4CH images the \textbf{HARD} dataset.

{\color{red} To test generalizablity across imaging modalities, we use a third dataset consisting of Cephalometric Radiographs, in which the images contain repetitive structures \cite{wang2016benchmark}. The dataset has a total of 19 annotated landmarks, where we use the junior annotator as the ground truth (following the convention of \cite{lindner2016fully, thaler2021modeling, zhong2019attention }). For our study of aleatoric uncertainty in Sec. \ref{section:aleatoric_uncertainty}, we use subset of 5 landmarks which have a total of 11 annotations provided by \cite{thaler2021modeling}. The images have a spatial resolution of $1935 \times 2400$ pixels, where each pixel represents 0.1mm of the structure. Fig. \ref{subfig:landmarkisbi} shows an example image annotated with the aleatoric uncertainty landmark subset.}

\vspace{-3mm}

\section{Experiments and Results}
\vspace{-1mm}
First, in Sec. \ref{section:baseline_results} we present the baseline landmark localization performance of PHD-Net and U-Net over both SA and 4CH datasets using the S-MHA, E-CPV, and E-MHA methods to extract coordinates. This gives us a comparison of the coordinate extraction performance from each of our methods, and a baseline to measure the effectiveness of each method's uncertainty estimation. Second, in Sec. \ref{section:quantile_bins_analysis} we interrogate how using Quantile Binning with our uncertainty measures delineates predictions in terms of their localization error, and compare the predicted bins to the ground truth error quantiles. We show a practical example of how filtering out highly uncertain predictions can dramatically increase the proportion of acceptable localization predictions. In Sec. \ref{section:error_bound_analysis} we assess how well the uncertainty measures can predict error bounds for each bin.  {\color{red}  Next, we demonstrate the generalizability of Quantile Binning in Sec. \ref{section:generalizability} on the more diverse Cephalometric dataset. In Sec. \ref{section:comparing_q_values}, we highlight the flexibility of the method by quantifying the effects of varying the number of quantile bins ($Q$) used. Finally, in Sec. \ref{section:aleatoric_uncertainty} we explore aleatoric uncertainty, demonstrating Quantile Binning's effectiveness on landmarks with high ambiguity, as well as sharing insights on our studied uncertainty measure's relationship with aleatoric uncertainty.} 
When comparing between $B_{1}, B_{2-4}, B_{5}$ we use an unpaired $t$-test ($p \leq 0.05$) to test for significance. When comparing uncertainty metrics among the same Bin category and model, we use a paired $t$-test  ($p \leq 0.05$) to test for significance.

\vspace{-4mm}

\subsection{Experimental Setup}
\vspace{-1mm}

\begin{table}
    \caption{Localization errors (mm) for the uncertainty methods outlined. \textit{All} indicates entire set of predictions; \textit{$B_{1}$} indicates subset with the \textit{lowest uncertainties}. Mean error and standard deviation are reported across all folds \& all landmarks. \textbf{Bold} indicates best results in row for the given dataset.}
    \vspace{-2mm}    

    \scriptsize
    \centering

    \setlength{\tabcolsep}{3pt}
    \begin{tabular}{l c c c c }

    \toprule
         &  \multicolumn{2}{c}{4 Chamber Images}  &  \multicolumn{2}{c}{Short Axis Images}  \\
    \cmidrule(lr){2-3}\cmidrule(lr){4-5}
      {Method}  & \multicolumn{1}{c}{U-Net}  & {PHD-Net}  & \multicolumn{1}{c}{U-Net} & {PHD-Net}  \\
   
    \cmidrule(lr){1-3}\cmidrule(lr){4-5}
         S-MHA All &\textbf{ 10.00 $\pm$ 18.99}  & 11.07 $\pm$  21.33 & 5.86 $\pm$ 14.19 & \textbf{3.58  $\pm$ 3.52} \\
        S-MHA $B_{1}$ & 6.79 $\pm$ 6.09  & \textbf{5.80 $\pm$ 9.03} & 3.62 $\pm$ 2.45 & \textbf{2.78  $\pm$ 1.99} \\
    \cmidrule(lr){1-3}\cmidrule(lr){4-5}

        E-MHA All &\textbf{ 6.36 $\pm$ 8.01} & 9.14 $\pm$ 18.11   & 4.37 $\pm$ 8.86  & \textbf{3.36 $\pm$ 3.50} \\
         E-MHA $B_{1}$ & 4.93 $\pm$ 2.85 & \textbf{4.70 $\pm$ 3.21}   & 2.98 $\pm$ 2.09  &\textbf{ 2.39 $\pm$ 1.90} \\
    \cmidrule(lr){1-3}\cmidrule(lr){4-5}

         E-CPV All & \textbf{8.13 $\pm$ 10.16} & 9.42 $\pm$ 13.07  & 4.97  $\pm$  7.51  & \textbf{3.22 $\pm$ 2.93}  \\
          E-CPV $B_{1}$ & 5.34 $\pm$ 3.00 & \textbf{5.10 $\pm$ 6.76}  & 3.75  $\pm$  2.13  & \textbf{2.47 $\pm$ 2.08 } \\
       
        \bottomrule
    \end{tabular}
    \label{table:allerrors}
     \vspace{-3mm}    

    \end{table}

We split both CMR datasets into 8 folds, and perform 8-fold cross validation for both U-Net and PHD-Net. {\color{blue} For each of the eight iterations, we select one fold as our testing set, one our hold-out validation set and the remaining 6 as our training set.} {\color{red} For the Cephalometric dataset we follow previous work (\hspace{1sp}\cite{lindner2016fully, thaler2021modeling, zhong2019attention}) and perform 4-fold cross validation using junior annotations, setting aside a random 20\% of each fold's training set as our hold-out validation-set. We resize the images to $512\times 512$} pixels and upsample the predicted heatmaps to the original image size before coordinate extraction. We select $T=5$ for the ensemble methods, training 5 identical, randomly initialized models at each iteration. We chose $T=5$ to compromise with computational constraints, asserting that 5 models are representative to compare the uncertainty methods for our purposes. {\color{blue}We randomly select a model from our trained ensemble for our S-MHA uncertainty measure. For our Quantile Binning method, we select $Q=5$ for 5 bins, striking a balance between the resolution of separation of the data and the limited size of our hold-out validation set ($\sim$30 samples for the CMR datasets, $\sim$60 samples for the Cephalometric dataset). We explore the effect of changing $Q$ in Sec. \ref{section:comparing_q_values}.}   

{\color{blue} We implement a vanilla U-Net model \cite{ronneberger2015u}.} We design the architecture with 5 encoding-decoding levels, creating 1.63M learnable parameters. {\color{red} Each level contains 2 residual units, where each residual unit applies a $3 \times 3$ convolution, instance normalization, and ReLU to the input, before concatenating the resulting output with the unit input. As we descend down the five levels of the encoder we use (16, 32, 64, 128, 256) input channels respective to each layer, mirroring this in the decoder path. On the encoder path we use maxpooling after each level to reduce spatial dimensions, and on the decoder path we use transposed convolutions to upsample the spatial resolution.} We modify the objective function from image segmentation to simultaneous landmark localization, minimising the mean squared error between the target and predicted heatmaps. We use the full $512\times 512$ pixel image as input, and learn heatmaps of the same size. We train for 1000 epochs with a batch size of 2, and a learning rate of 0.001 using the Adam Optimizer (settings from \cite{schobs2021confidence}).  {\color{blue} We generate target heatmaps using Eq. (\ref{equation:gaussian}) with a standard deviation of 8 for our CMR datasets and 2 for our Cephalometric dataset (chosen experimentally using the first fold of each dataset).} {\color{red} We do not use data augmentation.}

We implement our PHD-Net model following \cite{schobs2021confidence}, creating a model with 0.06M learnable parameters. For all experiments we trained PHD-Net for 1000 epochs using a batch size of 32 and a learning rate of 0.001, using the Adam Optimizer. We train one landmark at a time. Note, the only difference in setup from \cite{schobs2021confidence} in this work is different fold splits and training for an additional 500 epochs (same as U-Net) with no early stopping, since we now use our validation set for Quantile Binning. {\color{red} We do not use data augmentation.}

\vspace{-3.5mm}

\subsection{Baseline Landmark Localization Performance}

\label{section:baseline_results}
Table \ref{table:allerrors} shows the baseline performance for U-Net and PHD-Net at localizing landmarks in our 4CH and SA datasets. We make the following observations: 

\begin{itemize}

 \item When considering {\color{red} localization error} for the entire set of landmarks (\textit{All}), performance is better on the SA dataset for both models, with PHD-Net outperforming U-Net. On the 4CH dataset, U-Net outperforms PHD-Net {\color{red} in terms of fewer gross mispredictions}, suggesting the higher capacity model of U-Net is more robust to datasets with large variations. 

\item Simply using a single model with our S-MHA strategy is predictably less robust than ensemble approaches. 

\item E-MHA outperforms the previous strong baseline of E-CPV for coordinate extraction. However, does it outperform E-CPV in terms of uncertainty estimation? We explore this in Section \ref{section:quantile_bins_analysis}.
\item The standard deviation in the error for the baseline \textit{All} results in Table \ref{table:allerrors} is high for all models. We aspire to catch these bad predictions using Quantile Binning in Sec. \ref{section:quantile_bins_analysis}.

\end{itemize}
\vspace{-1mm}

\begin{figure*}[t]
    \centering
    \begin{subfigure}[t]{0.49\textwidth} 
         \centering
        \includegraphics[width=\linewidth]{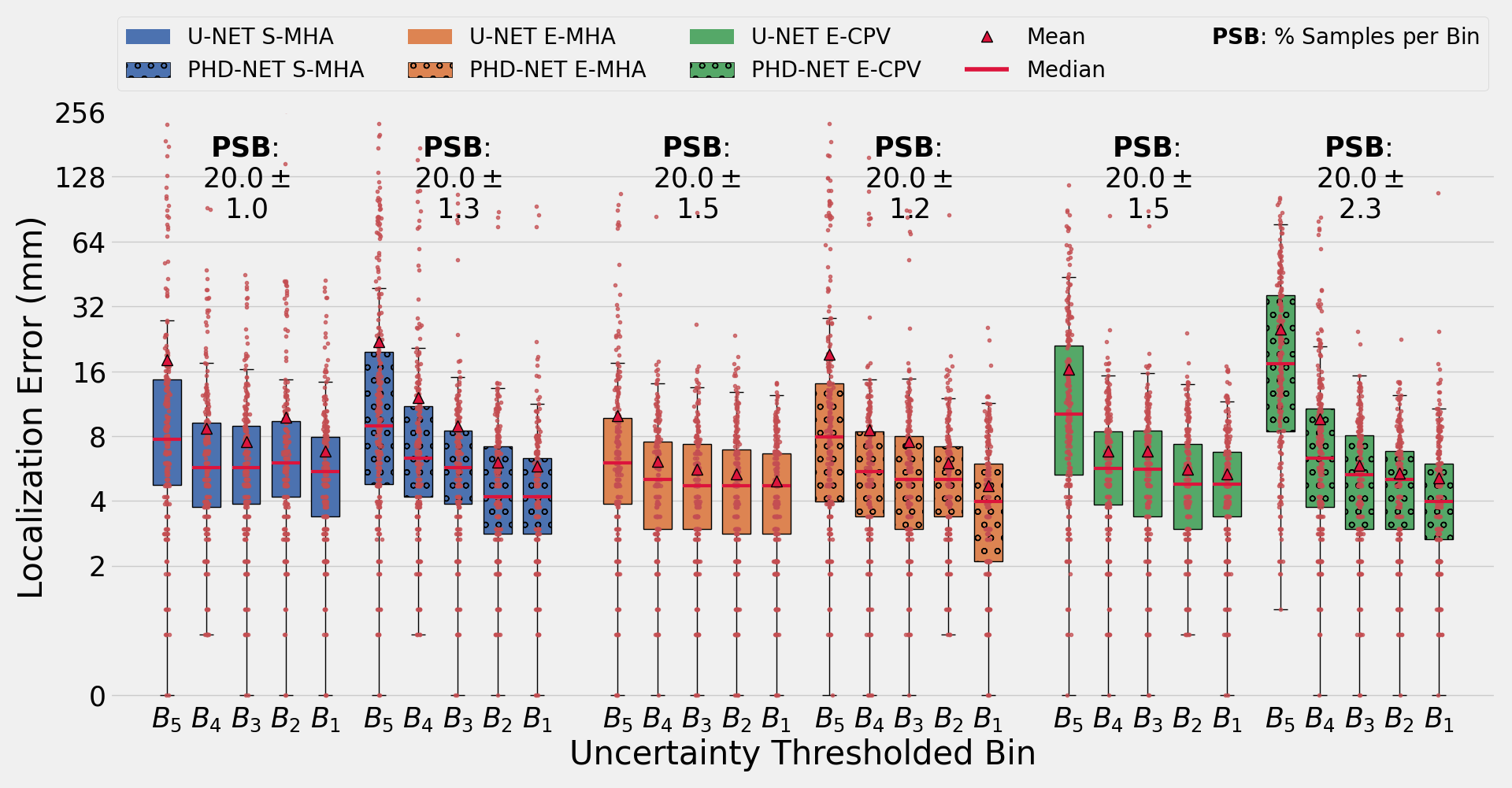}

        \caption{Localization error for each Bin - 4CH dataset (Lower is better).}
        \label{subfig:error_bins_4ch}
    \end{subfigure}
    \begin{subfigure}[t]{0.49\textwidth} %
          \centering
        \includegraphics[width=\linewidth]{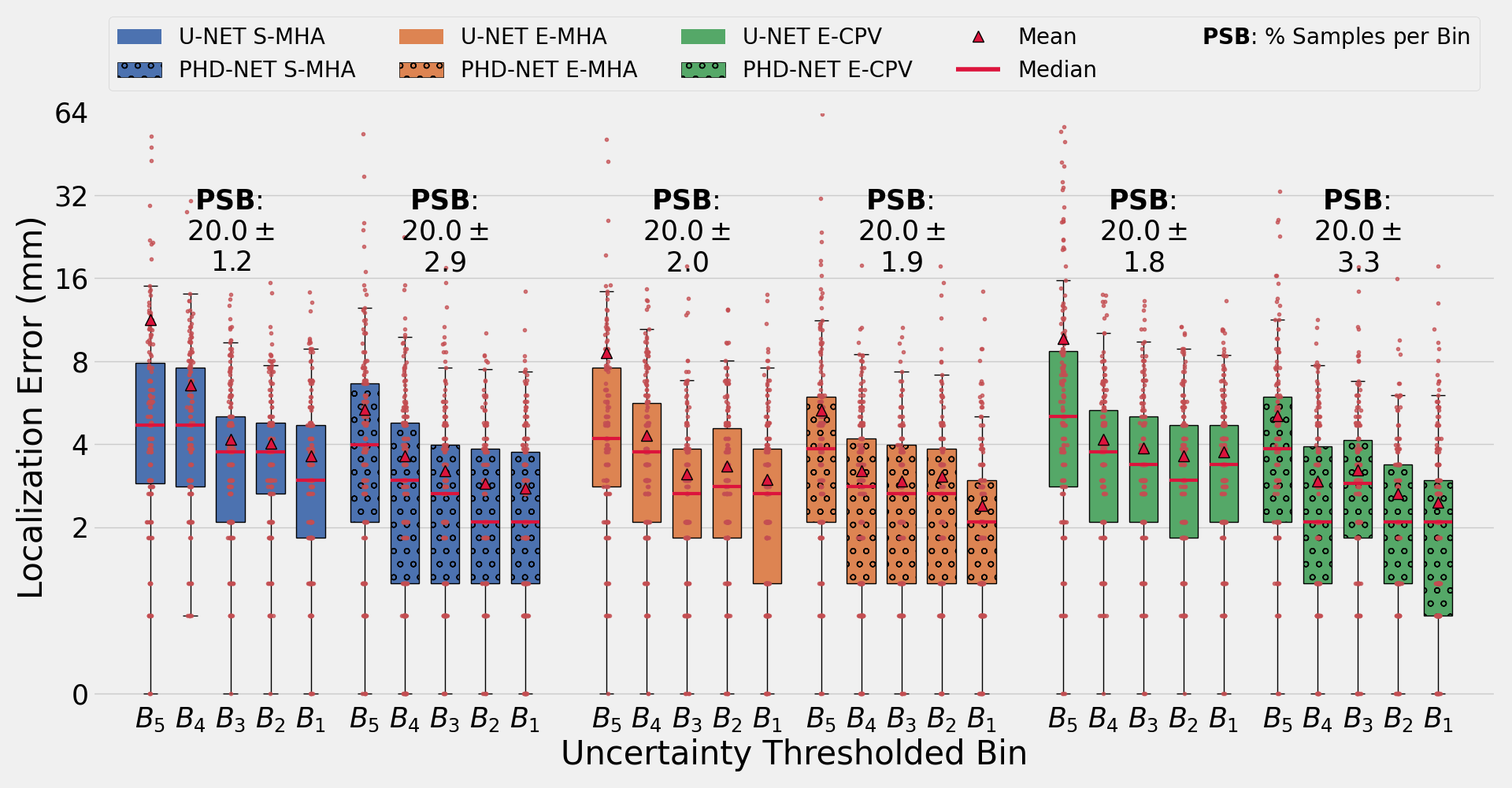}

    	\caption{Localization error for each Bin - SA dataset (Lower is better).}
    	\label{subfig:error_bins_sa}
    \end{subfigure} 
     \begin{subfigure}[b]{0.49\textwidth}
         \centering
        \includegraphics[width=\linewidth]{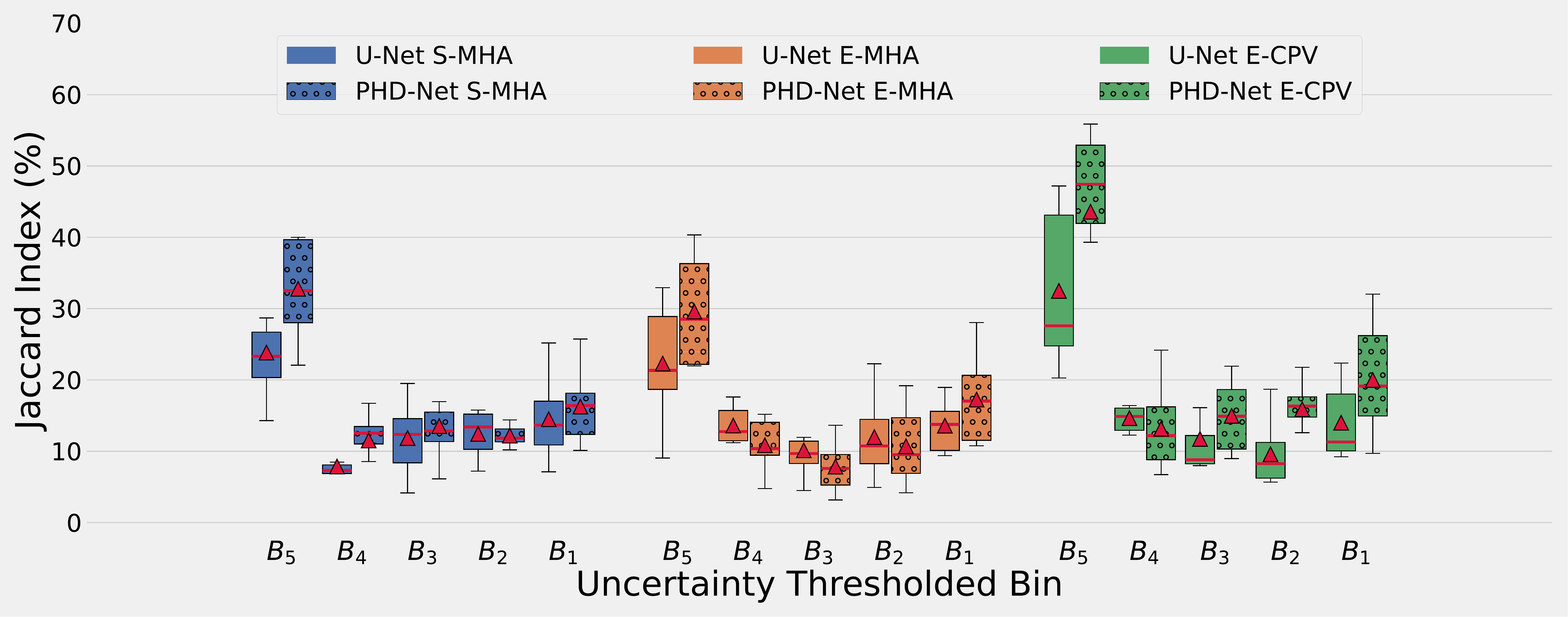}
        \caption{Jaccard Index for each Bin - 4CH dataset (Higher is better).}
    \label{subfigure:jaccard_4ch}
    \end{subfigure}
    \begin{subfigure}[b]{0.49\textwidth} %
        \centering
        \includegraphics[width=\linewidth]{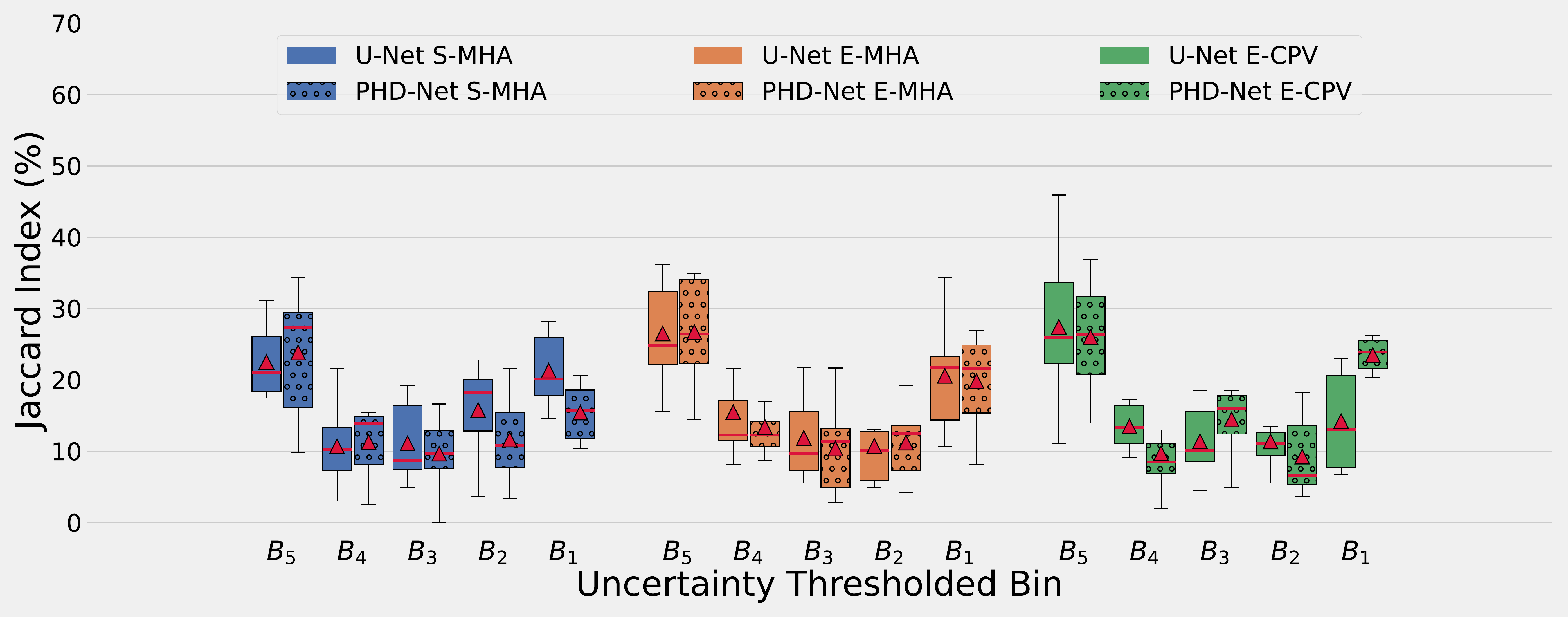}
    	\caption{Jaccard Index for each Bin - SA dataset (Higher is better).}
    	\label{subfigure:jaccard_sa}
    \end{subfigure}
    \begin{subfigure}[b]{0.49\textwidth} %
        \centering
    	\includegraphics[width=\linewidth]{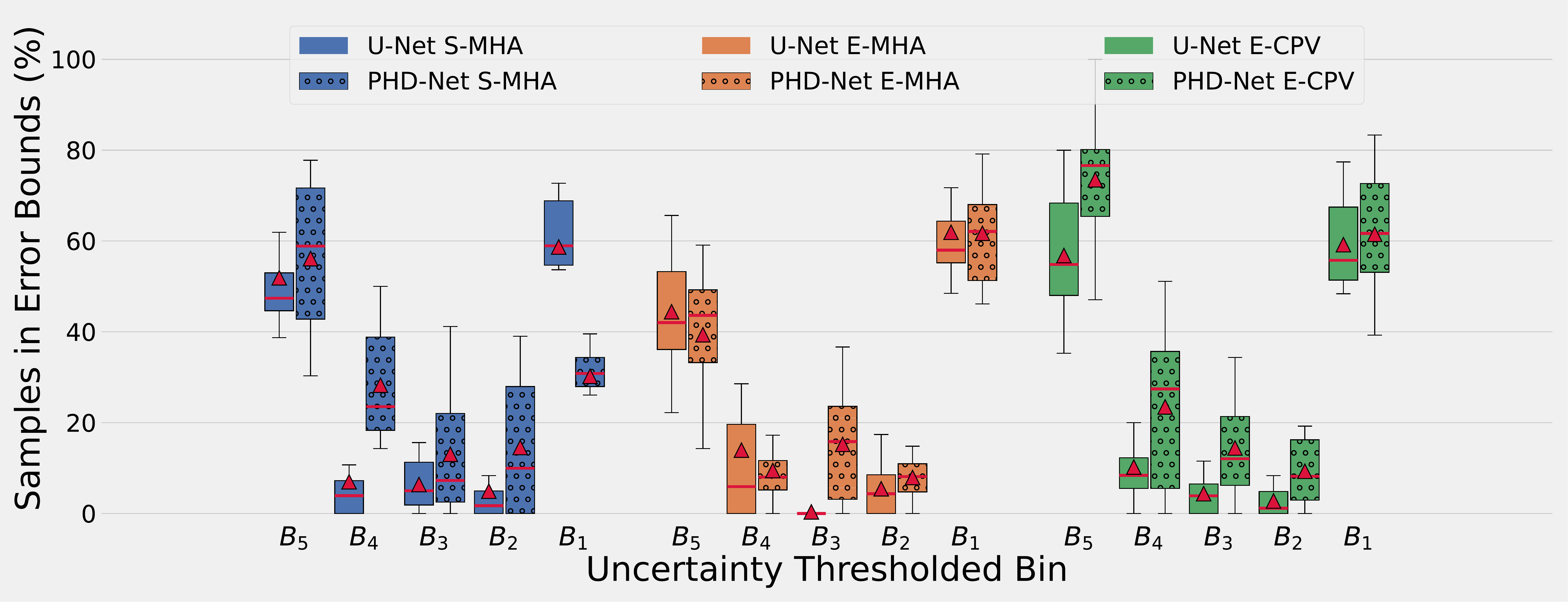}
    	\caption{Estimated Error Bound Accuracies- 4CH dataset (Higher is better).}
    	\label{subfig:errorbound4ch}
    \end{subfigure}
    \begin{subfigure}[b]{0.49\textwidth}
        \centering
        \includegraphics[width=\linewidth, ]{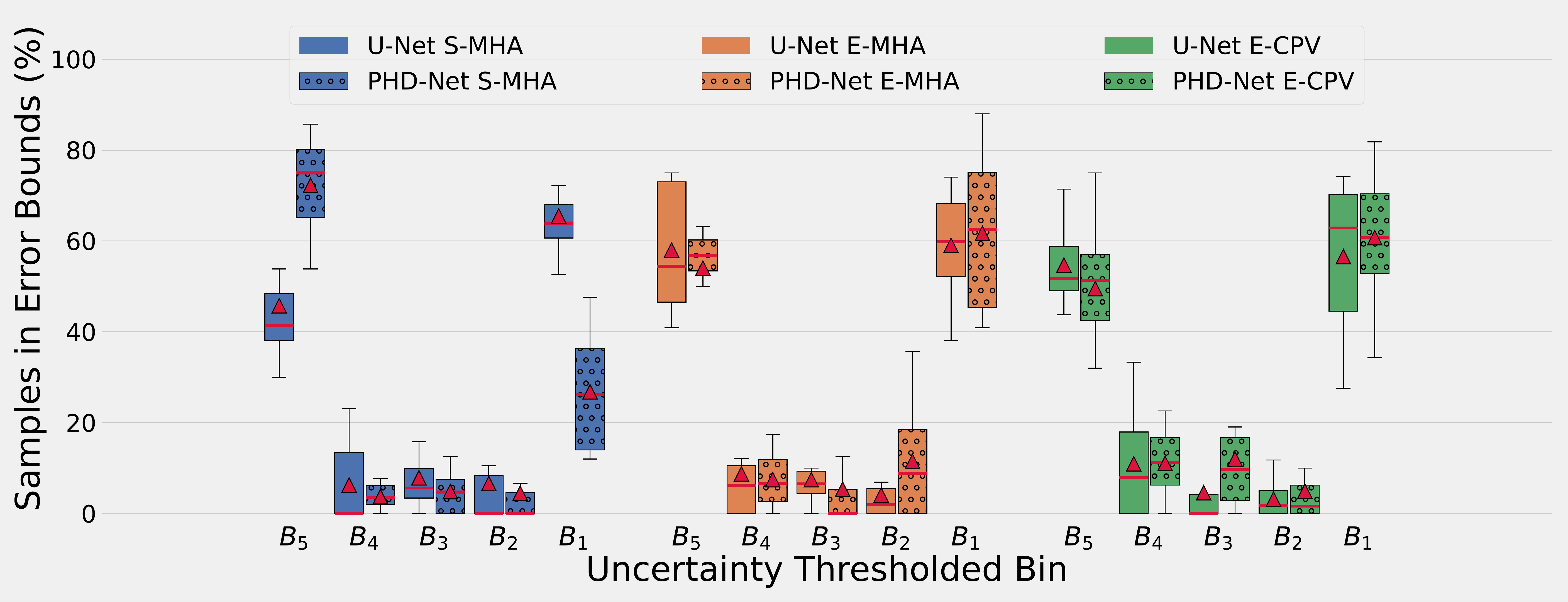}
        \caption{Estimated Error Bound Accuracies- SA dataset (Higher is better).}
        \label{subfig:errorboundsa}
    \end{subfigure}

    \caption{Results from Quantile Binning for U-Net and PHD-Net across all landmarks \& folds, using our three coordinate extraction \& uncertainty estimation methods. Bins are in descending order of uncertainty ($B_{5}$ highest uncertainty, $B_{1}$ lowest uncertainty). (a) and (b) show the mean localization error of each bin, with error decreasing as we move towards the bins with lower uncertainty. (c) and (d) present the Jaccard Index, showing how similar the predicted bins are to the ground truth error quantiles. (e) and (f) visualize the estimated error bound accuracy, showing the percentage of predictions within the estimated error bounds for each bin. Best viewed on screen.}
    \label{fig:all_quantile_results}
    \vspace{-6mm}

\end{figure*}

\vspace{-3mm}

\subsection{Analysis of the Predicted Quantile Bins}
\label{section:quantile_bins_analysis}
\vspace{-1mm}
We apply quantile binning to each uncertainty measure: S-MHA, E-MHA and E-CPV. We compare results over U-Net and PHD-Net for both the SA and 4CH datasets. 

{\color{red} First, we test our assumption that our uncertainty measures correlate with localization error. We report the Spearman's Rank Correlation Coefficient ($\rho$) since we are measuring a monotonic correlation which is not necessarily linear. All correlations are reported from the aggregated test sets across all eight folds of our CMR datasets, using a significance level of $p < 0.001$. For our 4CH dataset, S-MHA achieves correlations of 0.33 (weak-moderate)  \&  0.47 (moderate), E-MHA shows weak-moderate correlations of 0.39 \& 0.39, and E-CPV shows moderate correlations of 0.42 \& 0.53; for U-Net and PHD-Net respectively. For our SA dataset, S-MHA achieves correlations of 0.27 (weak)  \&  0.33 (weak-moderate), E-MHA weak-moderate correlations of 0.38 \& 0.38, and E-CPV correlations of 0.27 (weak), 0.36 (weak-moderate); for U-Net and PHD-Net, respectively. The correlation strength of S-MHA has high variance, whereas E-MHA shows a stable correlation across datasets and localization models. E-CPV achieves the strongest correlation with error across both models for our harder 4CH dataset, but a weaker correlation than E-MHA for our easier SA dataset. Overall, these results show that MHA and E-CPV contain information that can be exploited to estimate the uncertainty of our predictions.}

Next, we compare how our uncertainty measures can predict the true error quantiles. We found the most useful information is at the tail ends of the uncertainty distributions. Figs. \ref{subfigure:jaccard_4ch} \& \ref{subfigure:jaccard_sa} plot the Jaccard Index between ground truth error quantiles and predicted error quantiles. {\color{red} We notice a parabolic trend, where the outer bins are closer to the true error quantiles than the middle bins.} The highest uncertainty quantile bin ($B_{5}$) is significantly better at capturing the correct subset of predictions than the intermediate bins ($B_{2}- B_{4}$). Similarly, in {\color{red}some} cases the bin representing the lowest uncertainties ($B_{1}$) had a significantly higher Jaccard Index than the intermediate bins, but still lower than $B_{5}$. Figs. \ref{subfig:error_bins_4ch} \&  \ref{subfig:error_bins_sa} show the mean error {\color{red}($\blacktriangle$)} of the samples of each quantile bin over both datasets. {\color{blue} The most significant reduction in localization error is from $B_{5}$ to $B_{4}$ for all uncertainty measures. The sample distribution over the bins, indicated by the red dots, confirms that $B_{5}$ captures more gross mispredictions than the remaining bins, particularly for the 4CH dataset.} These findings suggest that most of the utility in the uncertainty measures investigated can be found at the tail ends of the scale. This is an intuitive finding, as the predictions in $B_{5}$ are \textit{certainly uncertain}, and the predictions in $B_{1}$ are \textit{certainly certain}. {\color{red} Figs. \ref{subfig:error_bins_4ch} \& \ref{subfig:error_bins_sa} show that each bin contains $\sim$20\% of the predictions, confirming our data-driven approach to setting uncertainty thresholds successfully approximates the true uncertainty distribution.}

The worse trained the landmark localization model, the more useful the uncertainty measure. Table \ref{table:allerrors} shows the localization error of all methods, models and datasets for the entire set (\textit{All}) and lowest uncertainty subset ($B_{1}$) of predictions. PHD-Net's baseline localization performance on the 4CH dataset was worse than U-Net. However, when we consider the lowest uncertainty subset of predictions ($B_{1}$), PHD-Net sees a 47\% average reduction in error from all predictions (\textit{All}), compared to U-Net's average reduction of 30\%. Similarly, U-Net performed worse than PHD-Net for the SA dataset, but saw an average error reduction of 31\% compared to PHD-Net's 25\%. This suggests that all investigated uncertainty measures are more effective at identifying gross mispredictions when models are poorly trained.

Using heatmap-based uncertainty measures is generalizable across heatmap generation approaches. The bin similarities in Figs. \ref{subfigure:jaccard_4ch} \& \ref{subfigure:jaccard_sa}  show that using S-MHA and E-MHA yields similar performance with PHD-Net  and U-Net, despite their different heatmap derivations. Surprisingly using E-MHA does not give a significant increase in bin similarity compared to S-MHA, suggesting the thresholds remain relatively stable across models.

No investigated method is conclusively best for estimating uncertainty in all scenarios. For the more challenging 4CH data, Fig. \ref{subfigure:jaccard_4ch} shows E-CPV is significantly better than S-MHA and E-MHA for both models at capturing the true error quantiles, corroborating the  findings of \cite{drevicky2020evaluating}. E-CPV is particularly good at identifying the worst predictions ($B_{5}$). For the easier SA data, no method has a significantly higher Jaccard Index. 
Therefore, when we generalize across both models and datasets, all uncertainty measures fared broadly similar on average in terms of error reduction between the entire set and the $B_{1}$ subset of predictions. S-MHA had an average error reduction of 35.07\%, E-MHA 32.94\% and E-CPV 32\%.

\begin{figure*}[t!]
    \begin{subfigure}[b]{0.24\linewidth} 
        \includegraphics[width=\linewidth]{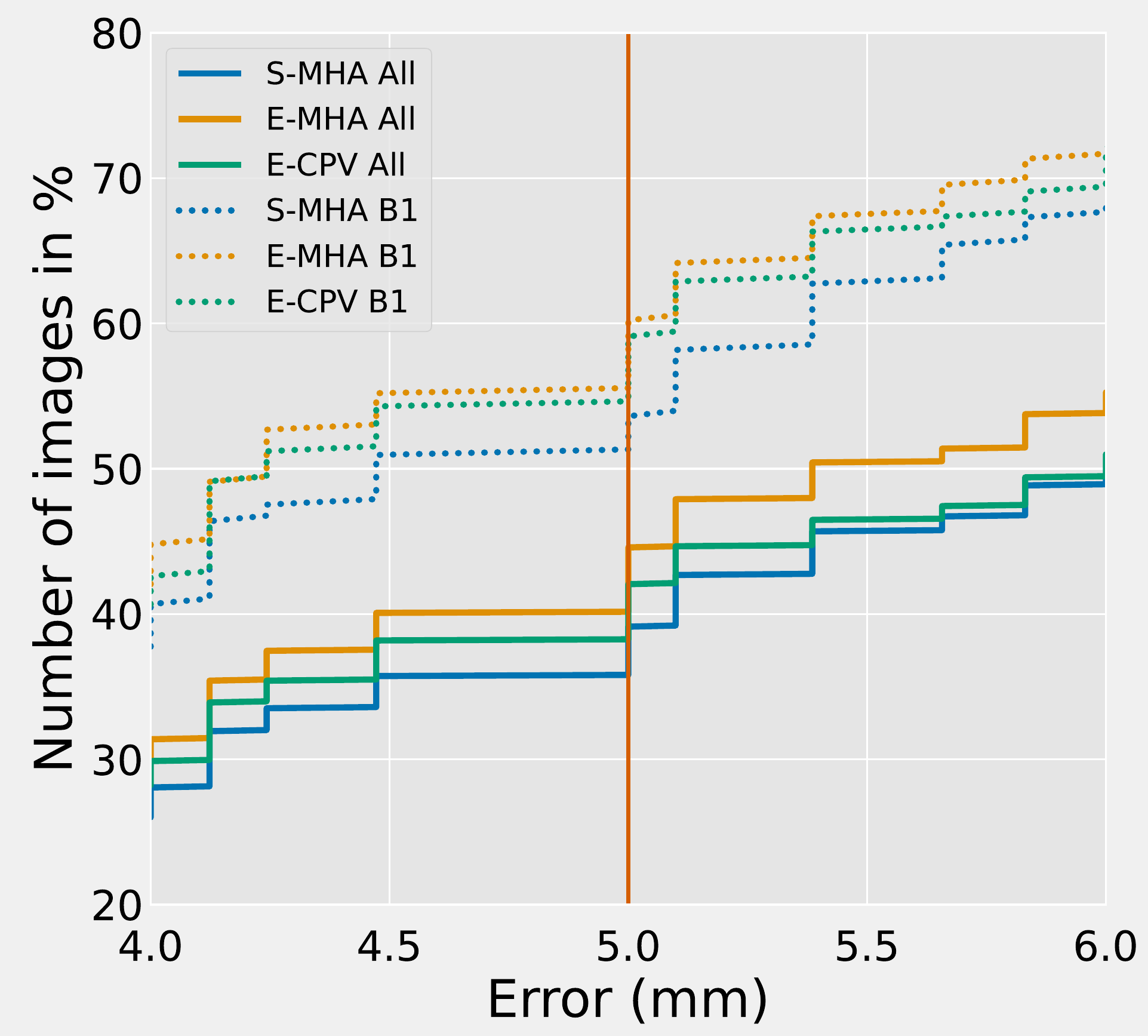}
        \caption{PHD-Net - 4CH Images.}
        \label{subfig: b4_error4chphdnet}
    \end{subfigure}
    \begin{subfigure}[b]{0.24\linewidth}
        \includegraphics[width=\linewidth]{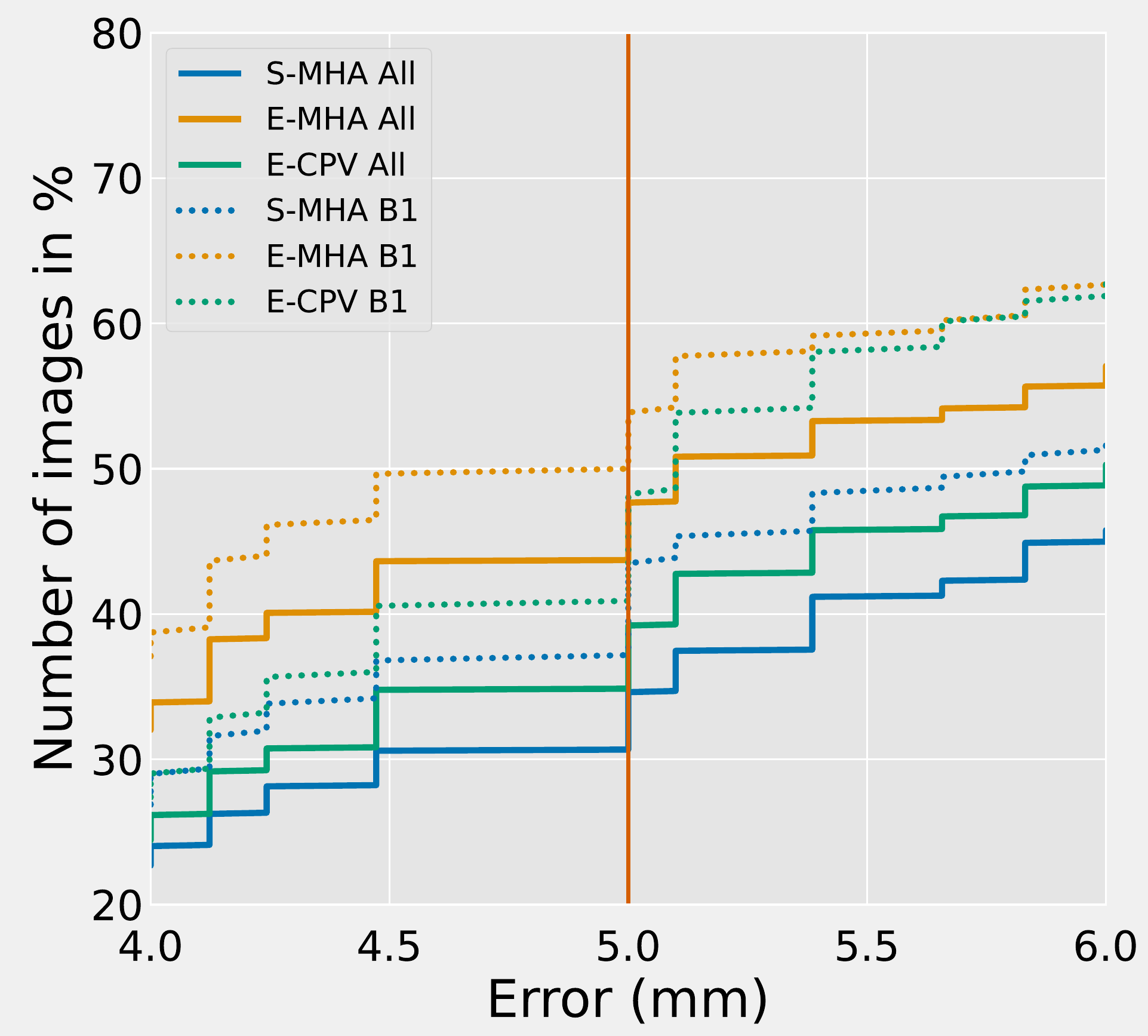}
        \caption{U-Net - 4CH Images.}
        \label{subfig: b4_error4chunet}
    \end{subfigure}
        \begin{subfigure}[b]{0.24\linewidth}
        \includegraphics[width=\linewidth ]{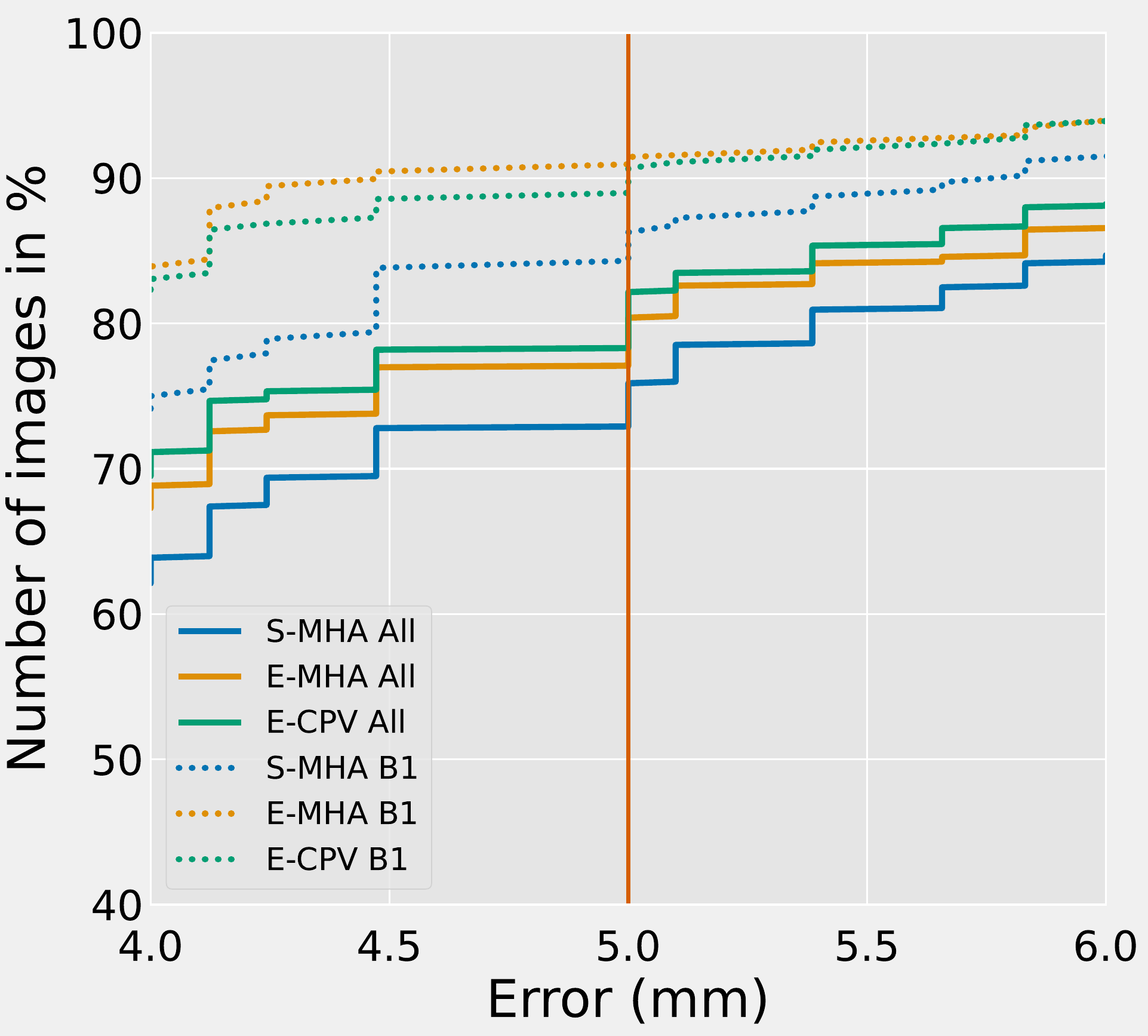}
        \caption{PHD-Net - SA Images.}
        \label{subfig: b4_errorsaphdnet}
    \end{subfigure}
    \begin{subfigure}[b]{0.24\linewidth}
        \includegraphics[width=\linewidth]{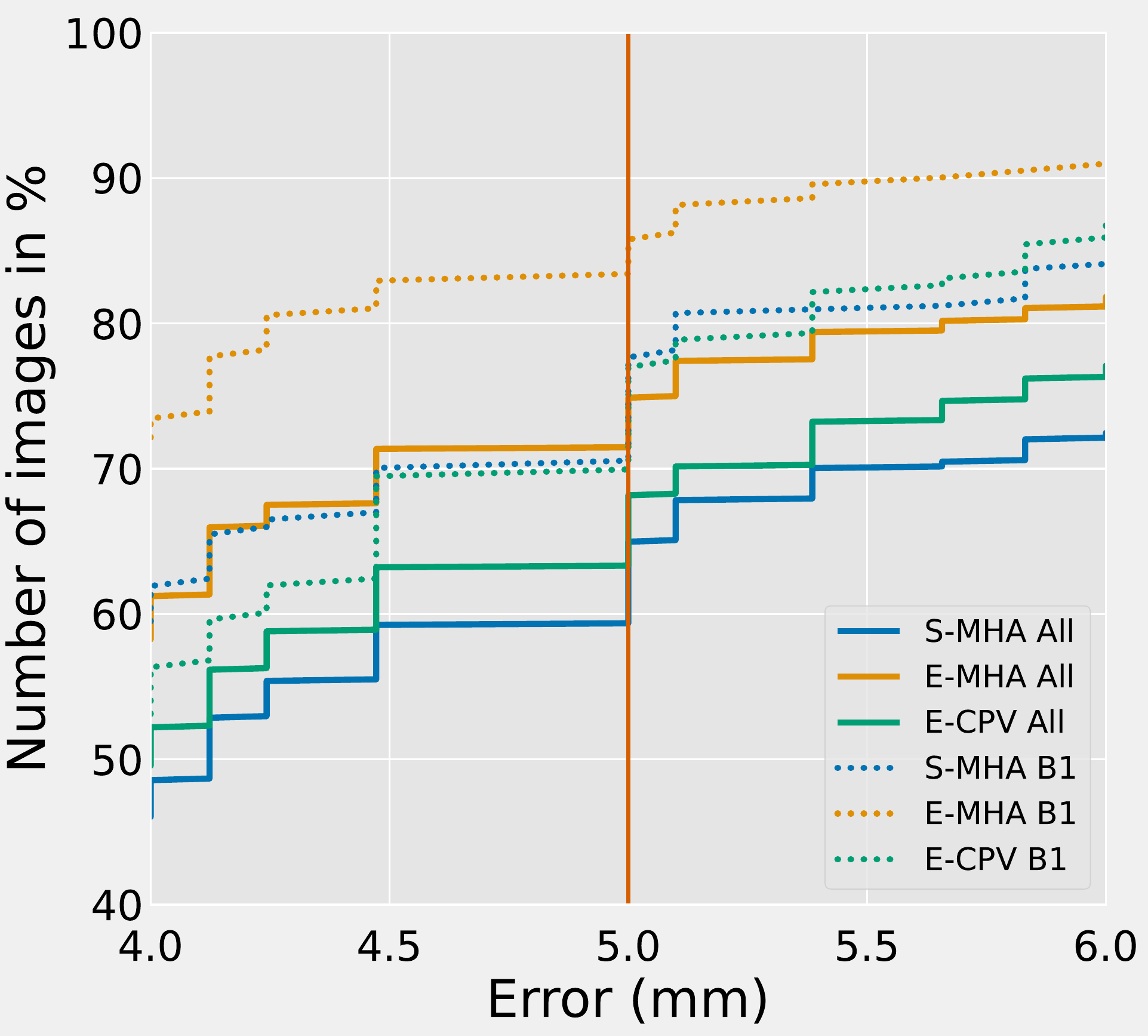}
        \caption{U-Net - SA Images.}
        \label{subfig: b4_errorsaunet}
    \end{subfigure}
        
\caption{Cumulative distribution of localization errors showing the \% of predictions under a given error threshold, comparing all predictions (\textit{All}) to the lowest uncertainty subset (\textit{$B_{1}$}) for the uncertainty methods across all folds \& landmarks. The vertical line is the acceptable error threshold, chosen by a radiologist. Higher percentage is better. }

\label{figure:all_errors}
\vspace{-6mm}
\end{figure*}

Despite similar performances in uncertainty estimation, we found E-MHA yields the greatest localization performance overall. Table \ref{table:allerrors} shows E-MHA offers the best localization performance for $B_{1}$ across both datasets and models. This is due to the combination of offering the most robust coordinate extraction on average (Table. \ref{table:allerrors}), and similar uncertainty estimation performance (Fig. \ref{subfigure:jaccard_4ch}, Fig. \ref{subfigure:jaccard_sa}). We more concretely demonstrate Quantile Binning's ability to identify low uncertainty predictions in Fig. \ref{figure:all_errors}. We clearly observe a significant increase in the percentage of images below the acceptable error threshold of 5mm when considering only predictions in $B_{1}$ - with E-MHA giving the greatest proportion of acceptable predictions.

\vspace{-2mm}

\vspace{-1mm}

\subsection{Analysis of Error Bound Estimation}
\label{section:error_bound_analysis}
\vspace{-1mm}

We analyse how accurate the isotonically regressed estimated error bounds  are for our quantile bins. Figs. \ref{subfig:errorbound4ch} \& \ref{subfig:errorboundsa} show the percentage of samples in each bin that fall between the estimated error bounds. 

We found we can predict the error bounds for the two extreme bins better than the intermediate bins. Figs. \ref{subfig:errorbound4ch} \& \ref{subfig:errorboundsa} show a similar {\color{red} parabolic} pattern to the Jaccard Index Figs. \ref{subfigure:jaccard_4ch} \& \ref{subfigure:jaccard_sa}, with the two extreme bins $B_{5}$ and $B_{1}$ predicting error bounds significantly more accurately than the inner bins. Again, this indicates the most useful uncertainty information is present at the extremes of the uncertainty distribution, with the predicted uncertainty-error function unable to capture a consistent relationship for the inner quantiles. {\color{blue} Further, the increased accuracy of the outer bins can be explained by the fact that it is easier to predict a single lower/upper bound than a pair of tighter bounds for the middling bins.} 

We also found that a well defined upper bound for heatmap activations is important for error bound estimates. For both the 4CH and SA datasets, S-MHA for PHD-Net is significantly more accurate at predicting error bounds for the highest uncertainty quantile $B_{5}$ compared to the lowest uncertainty quantile  $B_{1}$ (56\% \& 72\%  compared to 30\% \& 27\% for 4CH \& SA, respectively), correlating with S-MHA capturing a greater proportion of those bins (Jaccard Indexes of 32\% \& 24\% compared to 16\% \& 15\%). On the other hand, U-Net using S-MHA predicts error bounds for low uncertainty bins better than high uncertainty bins. This suggests that although PHD-Net's heatmap activation is a robust indicator of gross mispredictions, the upper error bound of $B_{1}$ ($\gamma_{1}$) cannot be accurately predicted due to the loose upper bound of the heatmap activations causing high variance. This is alleviated by using an ensemble of networks in E-MHA, where the $B_{1}$ bound accuracy is improved to 62\%.

E-MHA and E-CPV are more consistent than S-MHA. Overall, there is no significant difference between the error bound estimation accuracy of E-MHA and S-MHA, but Figs. \ref{subfig:errorbound4ch} \& \ref{subfig:errorboundsa} show E-MHA has less variation in performance between U-Net and PHD-Net compared to S-MHA, suggesting an ensemble of models is more robust. For the 4CH dataset, PHD-Net using E-CPV is on average significantly more accurate at predicting error bounds than S-MHA and E-MHA. However, there are no significant differences for PHD-Net on the easier SA dataset, nor U-Net on either dataset. There are also no significant differences between U-Net and PHD-Net in error bound estimation accuracy, with each method broadly equally effective for both models.

\begin{figure*}[!ht]
    \centering
    \begin{subfigure}[t]{0.49\textwidth} 
         \centering
         \includegraphics[width=\linewidth]{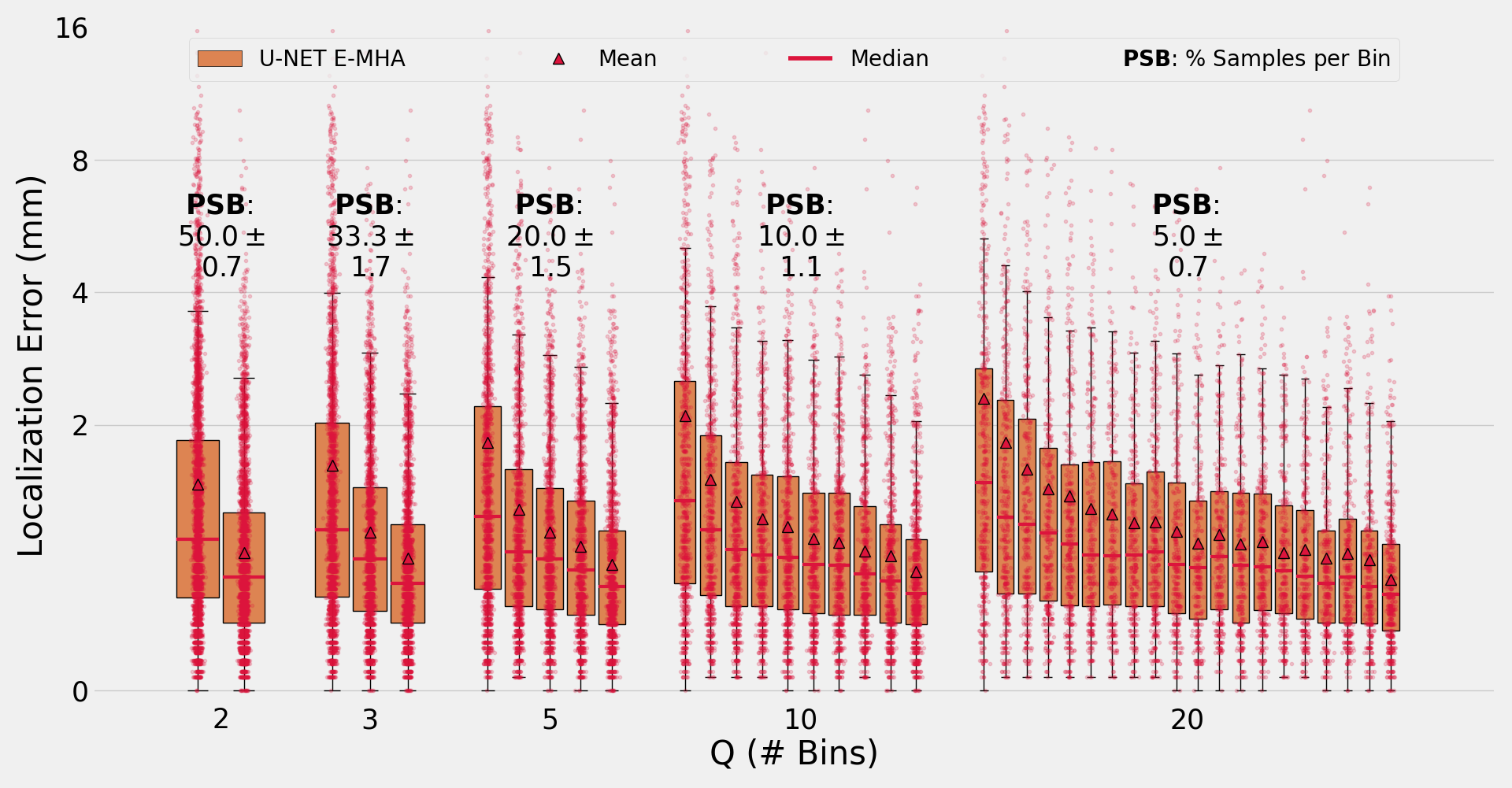}

        \caption{Localization error - E-MHA (Lower is better).}
        \label{subfig:error_bins_isbi_emha}
    \end{subfigure}
    \begin{subfigure}[t]{0.49\textwidth} %
          \centering
         \includegraphics[width=\linewidth]{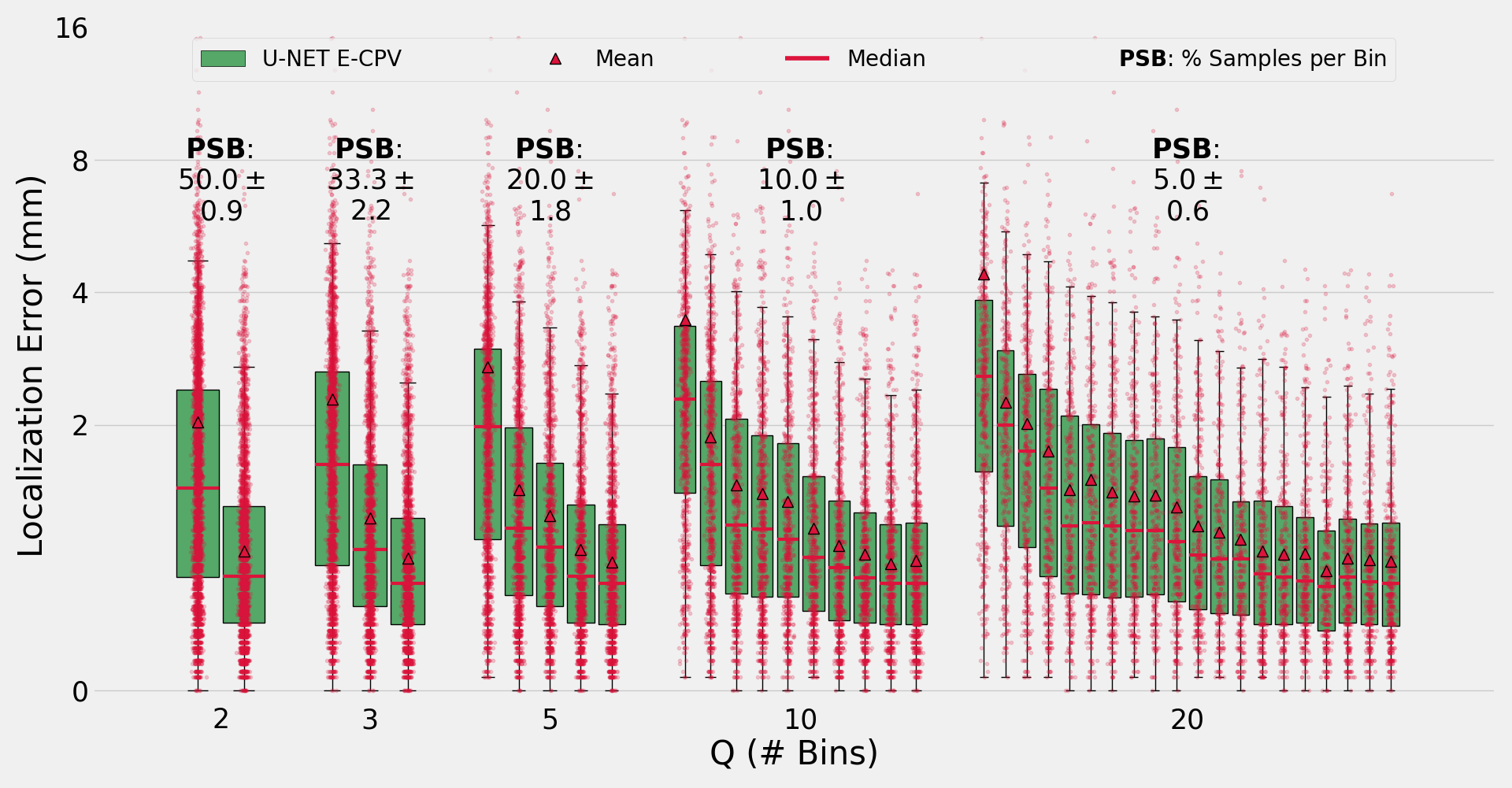}

    	\caption{Localization error - E-CPV (Lower is better).}
    	\label{subfig:error_bins_isbi_ecpv}
    \end{subfigure} 
     \begin{subfigure}[b]{0.49\textwidth}
         \centering
         \includegraphics[width=\linewidth]{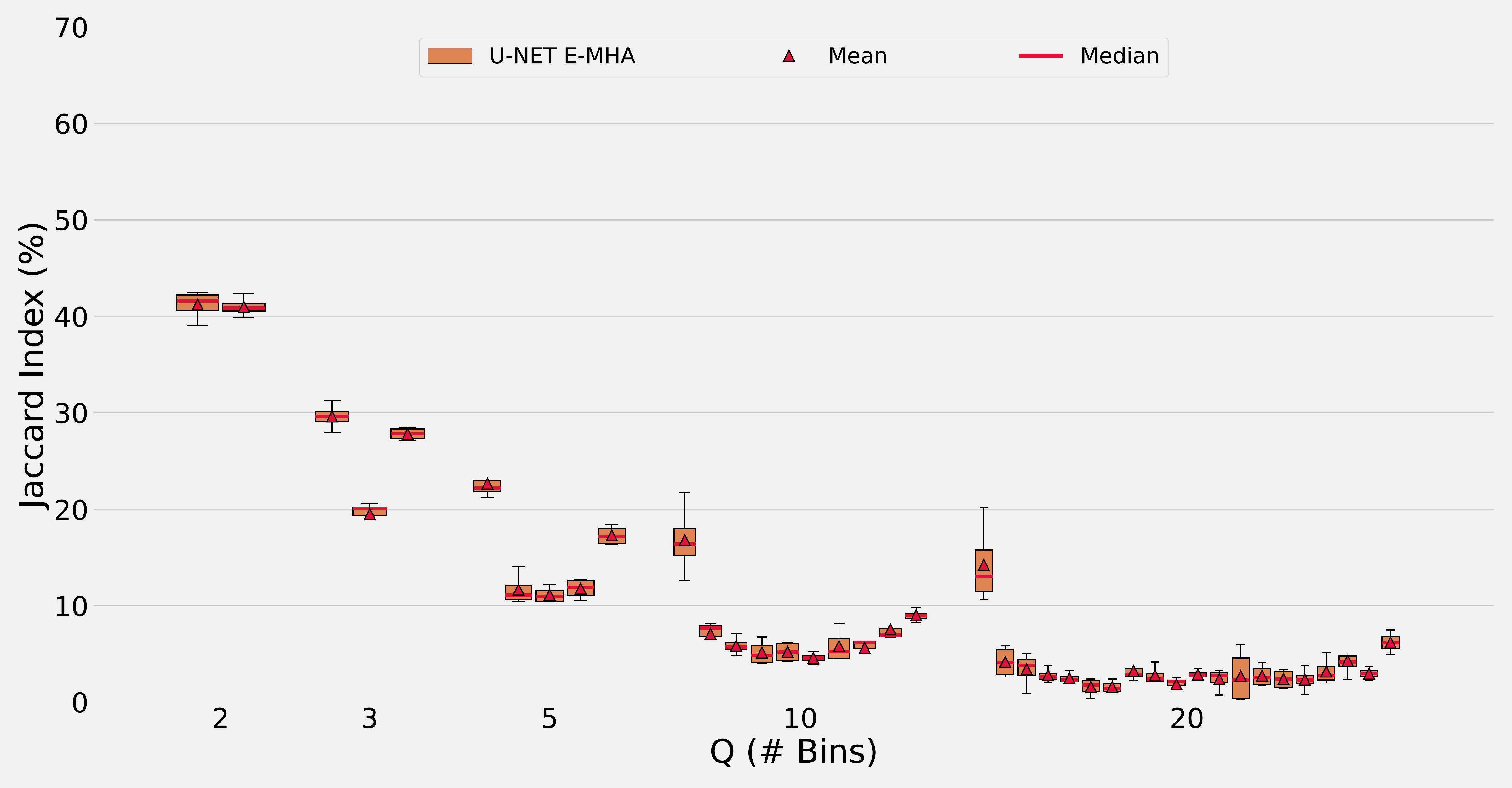}

        \caption{Jaccard Index - E-MHA (Higher is better).}
    \label{subfigure:jaccard_isbi_emha}
    \end{subfigure}
    \begin{subfigure}[b]{0.49\textwidth} %
        \centering
         \includegraphics[width=\linewidth]{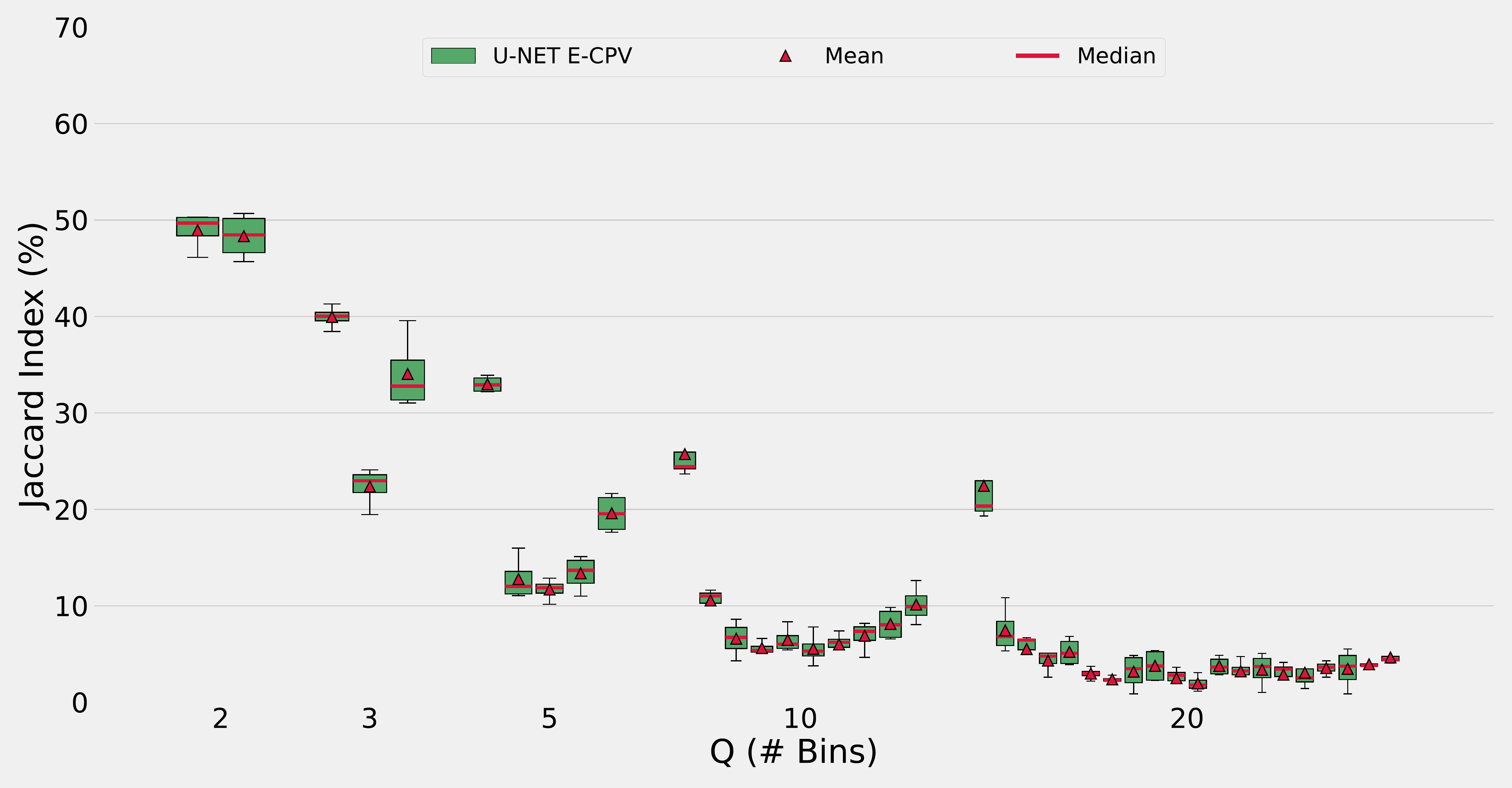}

    	\caption{Jaccard Index - E-CPV (Higher is better).}
    	\label{subfigure:jaccard_isbi_ecpv}
    \end{subfigure}
    \begin{subfigure}[b]{0.49\textwidth} %
        \centering
    	\includegraphics[width=\linewidth]{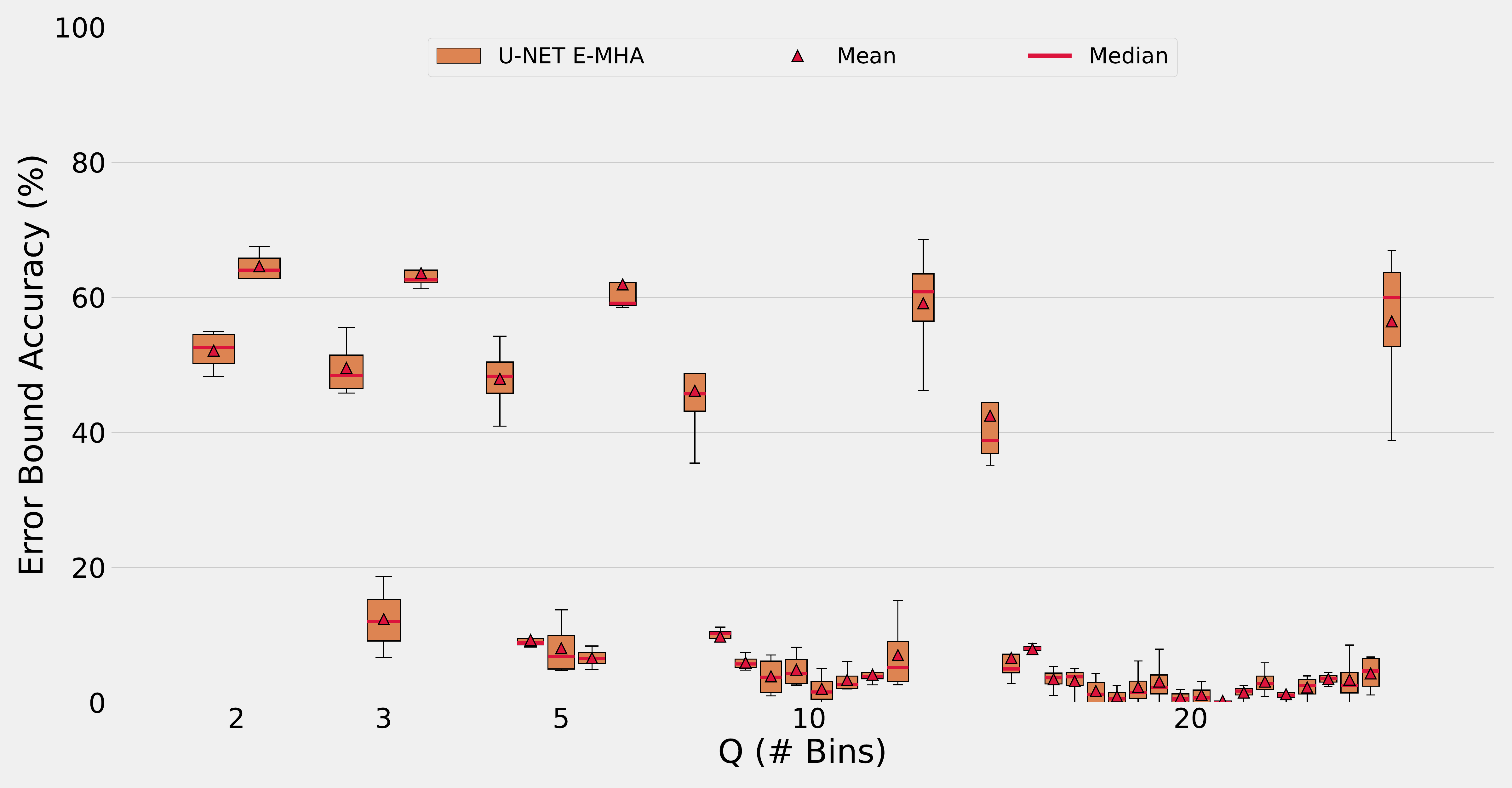}
    	\caption{Estimated Error Bound Accuracies - E-MHA (Higher is better).}
    	\label{subfig:errorbounds_isbi_emha}
    \end{subfigure}
    \begin{subfigure}[b]{0.49\textwidth}
        \centering
    	\includegraphics[width=\linewidth]{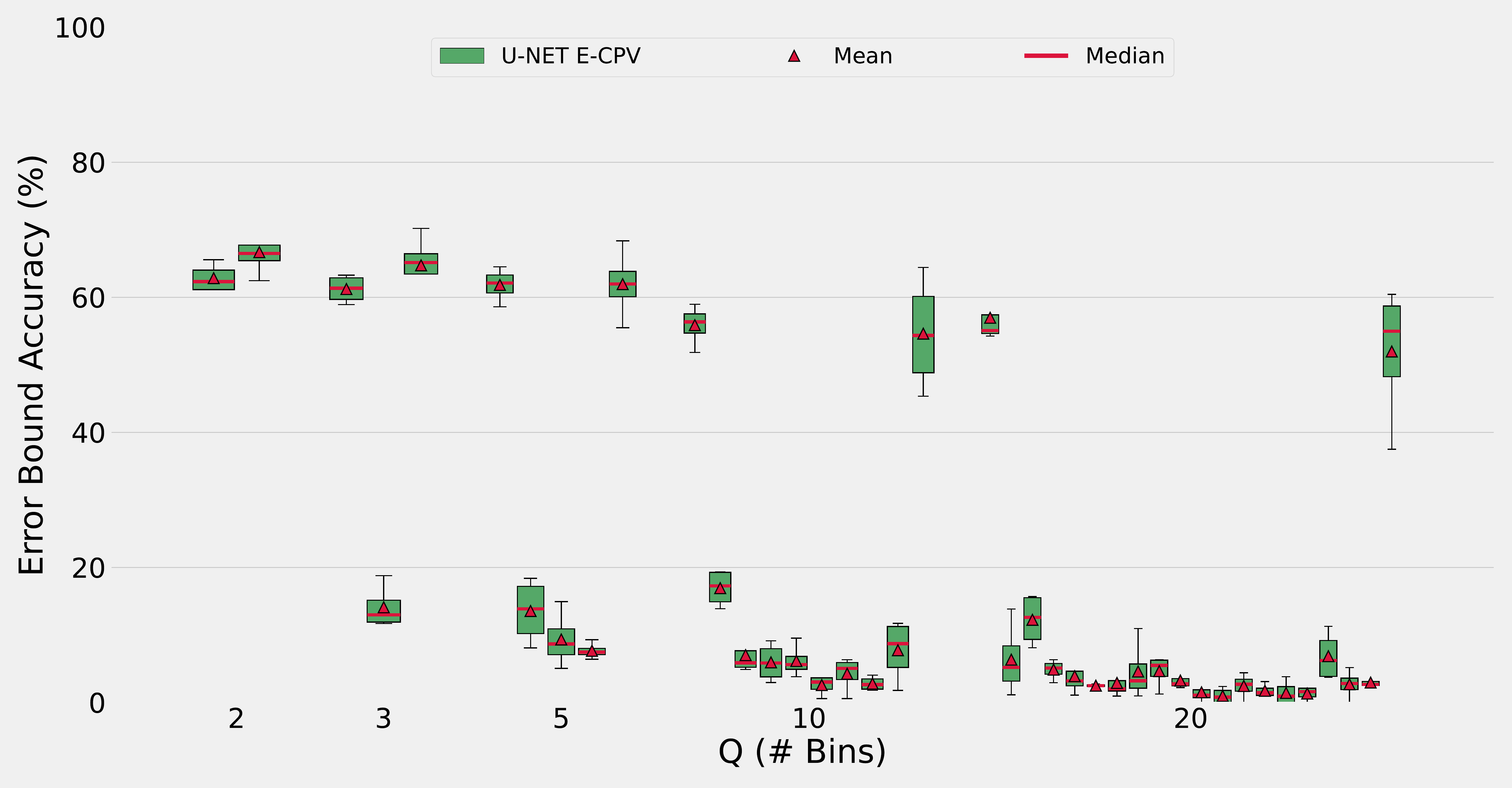}

        \caption{Estimated Error Bound Accuracies - E-CPV (Higher is better).}
        \label{subfig:errorbounds_isbi_ecpv}
    \end{subfigure}

     \caption{{\color{red}Quantile Binning varying $Q$ (Number of Quantile Bins) on the Cephalometric dataset. We show results for the uncertainty measures E-MHA and E-CPV, over all landmarks from a 4-fold CV, trained on the U-Net model. Red dots represent the errors of individual samples, best viewed on screen.}}
    \label{fig:comparing_q_emha_ecpv_isbi}
    \vspace{-6mm}

\end{figure*}

\vspace{-3mm}

{\color{red}

\subsection{Generalizability}
\label{section:generalizability}
\vspace{-1mm}
We train U-Net on the Cephalometric dataset, applying Quantile Binning to E-MHA and E-CPV to test their generalizability across imaging modalities. For $Q=5$, Figs. \ref{subfigure:jaccard_isbi_emha} \& \ref{subfigure:jaccard_isbi_ecpv} show a predictive power of true error quantiles comparable with the CMR datasets. The mean Jaccard Index (JI) for $B_{5}$ is 22\% for E-MHA and 34\% for E-CPV on the Cephalometric dataset, compared to 22\% \& 32\% for U-Net on the 4CH dataset. $B_{1}$ shows a better result than the CMR datasets, achieving a JI of 18\% for E-MHA and 19\% for E-CPV, compared to 15\% \& 14\% for U-Net on the 4CH dataset. E-CPV more effectively identifies the extreme mis-predictions compared to E-MHA, as evidenced by a higher JI for $B_{5}$ (left-most bin) in Figs. \ref{subfigure:jaccard_isbi_emha} \&  \ref{subfigure:jaccard_isbi_ecpv}, supporting the results from the challenging 4CH dataset. For $Q=5$, Figs. \ref{subfig:error_bins_isbi_emha} and \ref{subfig:error_bins_isbi_ecpv} show a gradual reduction in error from $B_{5}$ (left-most) to $B_{1}$. Overall, the larger Cephalometric dataset (19 landmarks) shows a more consistent downward trend in error across bins compared to our smaller CMR datasets (3 landmarks).

Next, to test the robustness of using  MHA as an uncertainty measure across target heatmaps of varying sizes, we repeated these experiments changing the standard deviation of the target heatmap from Eq. (\ref{equation:gaussian}) to 2, 4, 8 and 12. We found the trends of our Quantile Binning results hold, with only the localization error deteriorating as we increased the size of the Gaussian. We conclude that as long as the standard deviation leads to a learnable heatmap, similar uncertainty estimation properties are exhibited by MHA.

\vspace{-3mm}
\subsection{Varying Quantile Binning Resolution}
\label{section:comparing_q_values}
\vspace{-1mm}
We vary the number of Quantile Bins ($Q = \{2,3,5,10,20\}$) for the larger Cephalometric dataset to gain deeper insights on the flexibility of Quantile Binning. Figs. \ref{subfig:error_bins_isbi_emha} \& \ref{subfig:error_bins_isbi_ecpv} show the localization error quantiles across $Q$ for the Cephalometric dataset, with a gradual reduction in mean localization error ($\blacktriangle$) from $B_{Q}$ to $B_{1}$ for all values of $Q$. We find that the edge bins are most useful for all values of $Q$, with the Jaccard Indexes in Figs. \ref{subfigure:jaccard_isbi_emha} \& \ref{subfigure:jaccard_isbi_ecpv} and error bound accuracies in Figs. \ref{subfig:errorbounds_isbi_emha} \& \ref{subfig:errorbounds_isbi_ecpv} showing parabolic trends, confirming our results from the CMR datasets.

Further, Quantile Binning provides utility for a range of $Q$ values. First, consider the extreme case of $Q=2$, where the threshold is the median uncertainty of the validation set. Here, $B_{2}$ (the high uncertainty bin, left) captures the majority of the gross mispredictions and $B_{1}$ (the low uncertainty bin, right) captures the majority of the best predictions. Now, consider the effect of increasing $Q$, shown in Figs. \ref{subfig:error_bins_isbi_emha} \& \ref{subfig:error_bins_isbi_ecpv}. As we increase the number of Quantile Bins, the mean error ($\blacktriangle$) of $B_{Q}$ (far left bin of each set) increases. This is because as $Q$ increases, $B_{Q}$ is pushed farther towards the edge of the uncertainty measure distribution, capturing progressively more extreme outliers. Therefore, as $Q$ increases, we observe an increasingly logarithmic trend of the mean error across the bins as poor predictions are filtered out more gradually. 

In practice, the higher the value of $Q$, the greater the resolution of separation of the data. For example, consider the task of flagging up uncertain landmark predictions for manual review. Using $Q=2$ and flagging predictions from the highest uncertainty bin will lead to 50\% of predictions requiring review and re-annotation. On the other hand, filtering out the highest uncertainty bin using $Q=10$ leaves only 10\% of predictions to be reviewed. In each case, the user will have an upper error bound estimate for the remaining predictions with reasonable accuracy ($\sim$50\% for E-MHA and $\sim$60\% for E-CPV, the left-most bins, $B_{Q}$, in Figs. \ref{subfig:errorbounds_isbi_emha} \& \ref{subfig:errorbounds_isbi_ecpv}). However, the contents of $B_{Q}$ are more accurate when $Q$ is small, with a Jaccard Index of 50\% for $Q=2$ compared to 25\% for $Q=10$ for E-CPV (Fig. \ref{subfigure:jaccard_isbi_ecpv}). Therefore, this trade-off between true error quantile accuracy and binning resolution means $Q$ is a subjective choice that depends on the specificity of the downstream task and the resources available for reannotation.

Similar trends are present for our 4CH dataset and SA dataset, but we note that results are poor for $Q>=10$ compared to the Cephalometric dataset. This is because when fitting the data for Quantile Binning, our CMR datasets had access to a much smaller validation set compared to the Cephalometric dataset ($\sim$30 samples compared to $\sim$60 samples) and could not accurately estimate the quantile uncertainty distribution for large values of $Q$. Therefore, the larger the available validation set, the larger $Q$ can be set.

}
\vspace{-1mm}
\begin{figure*}[t]

 \begin{tabular}{M{0.1cm} M{2cm} M{7cm} M{7cm}}
 
    & Annotator Dist. & Quantile Errors & Jaccard Index \\
    \rotatebox[origin=c]{90}{$L_{4}$} & 
    \includegraphics[width=\linewidth]{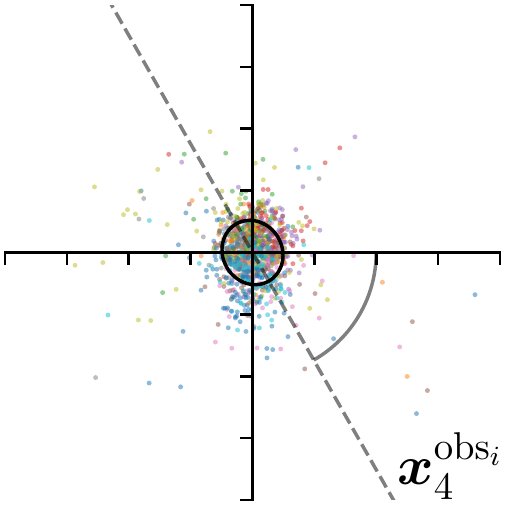} &  
    \includegraphics[width=\linewidth]{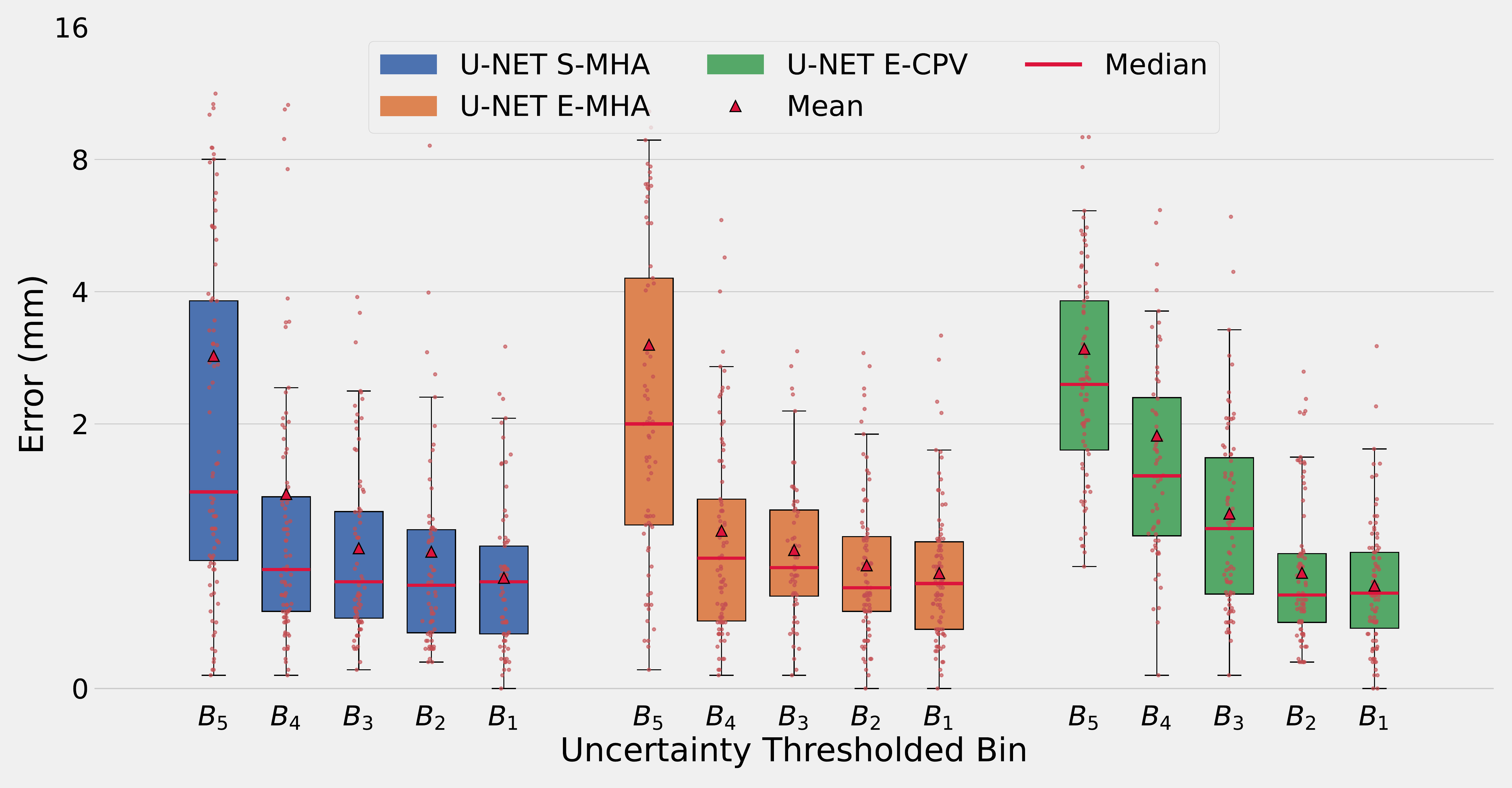} &  
    \includegraphics[width=\linewidth]{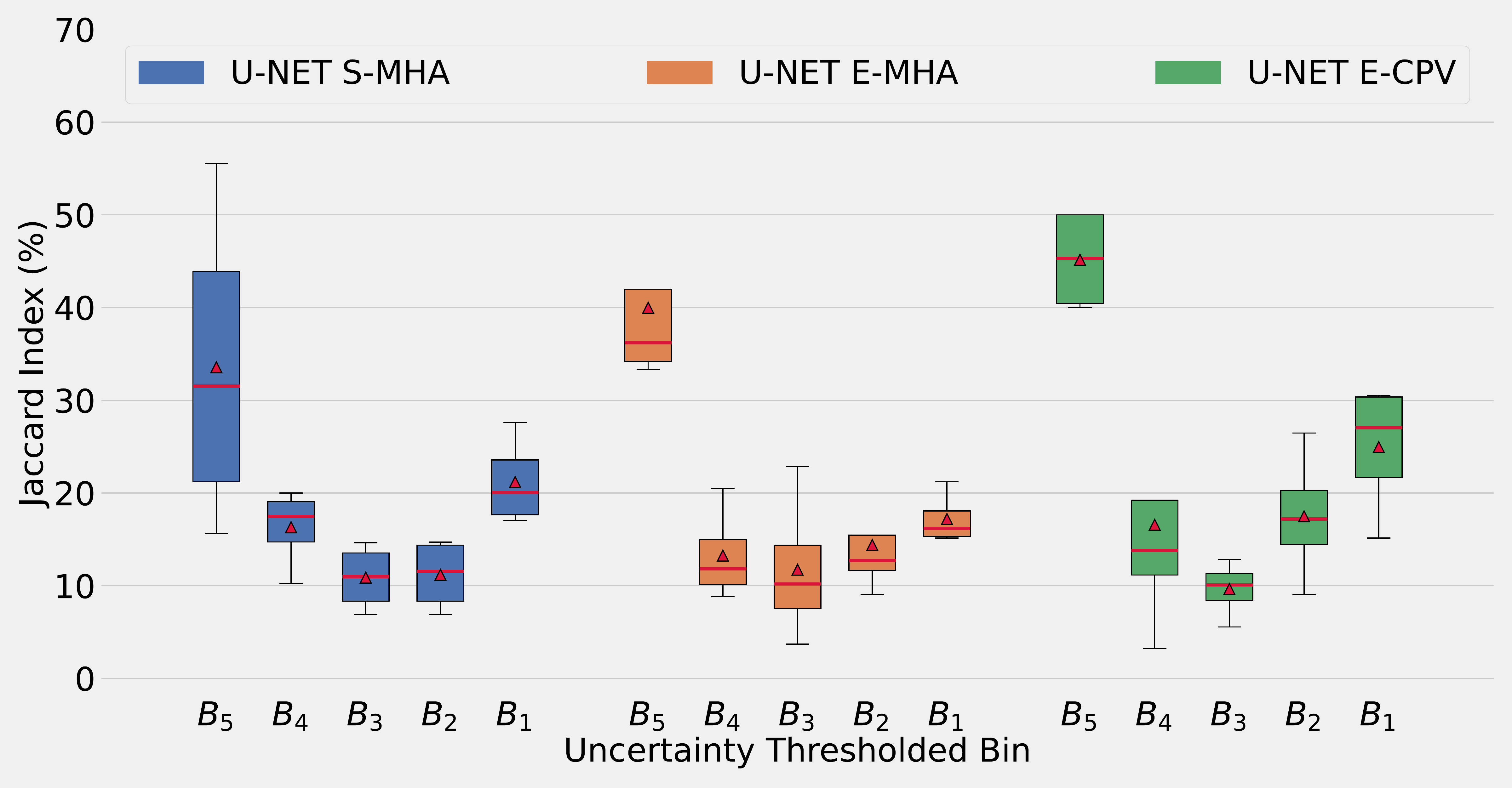} \\

    \rotatebox[origin=c]{90}{$L_{1}$} &
    \includegraphics[width=\linewidth]{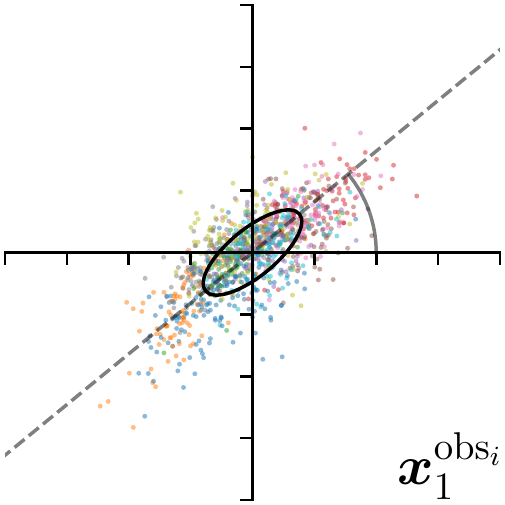} & 
    \includegraphics[width=\linewidth]{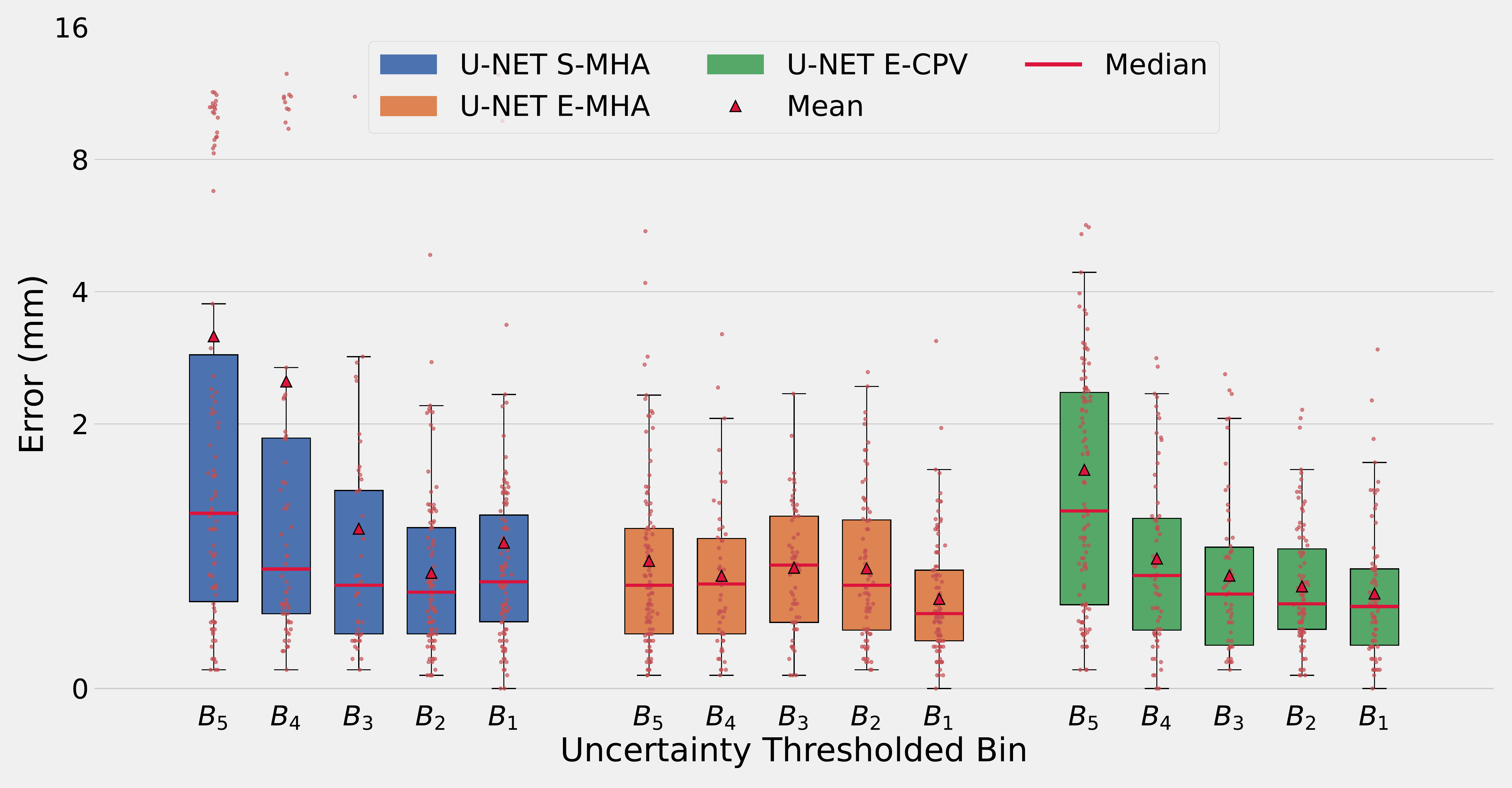}  & 
    \includegraphics[width=\linewidth]{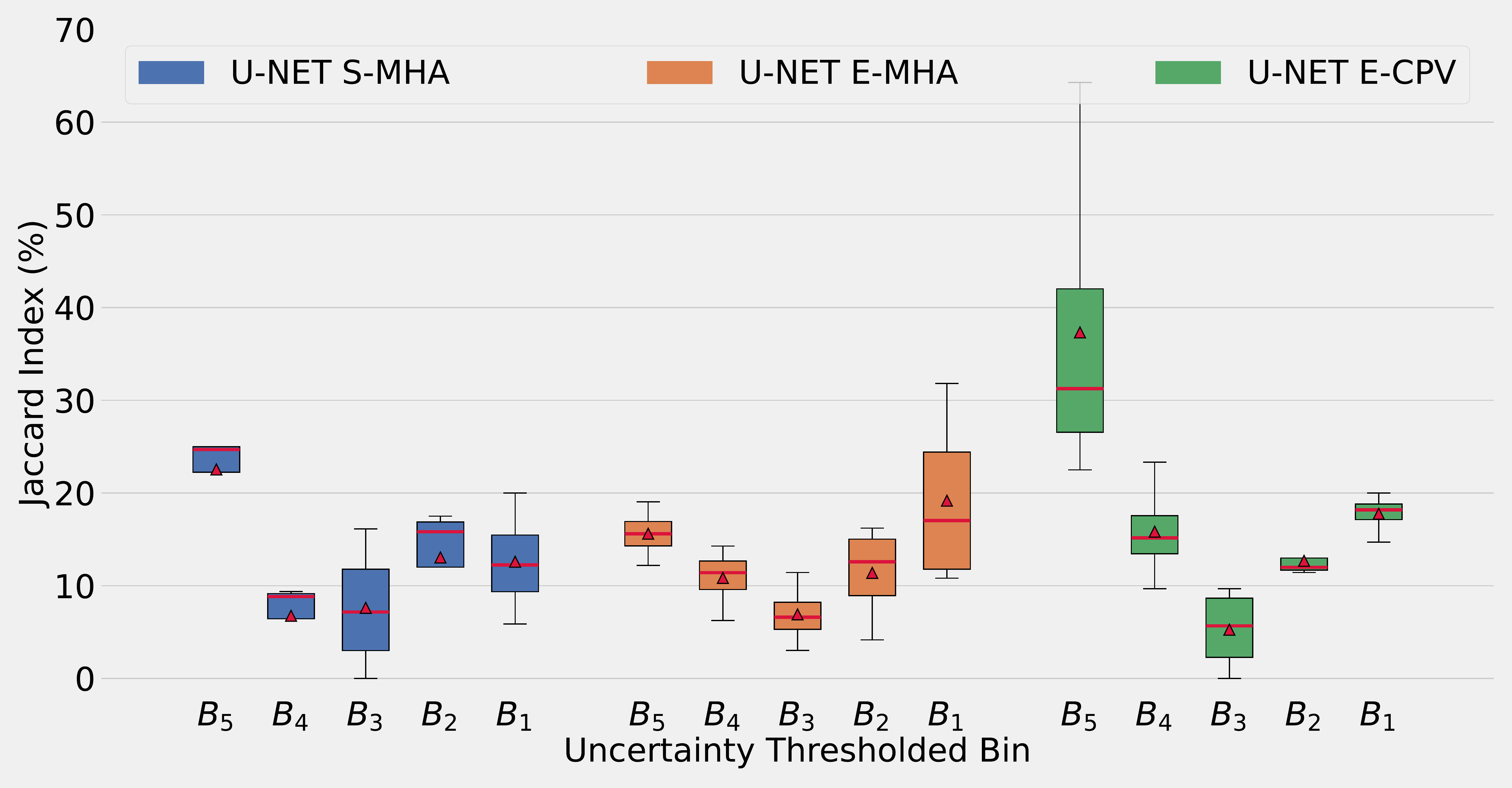}  \\
    
    \rotatebox[origin=c]{90}{$L_{5}$} & 
    \includegraphics[width=\linewidth]{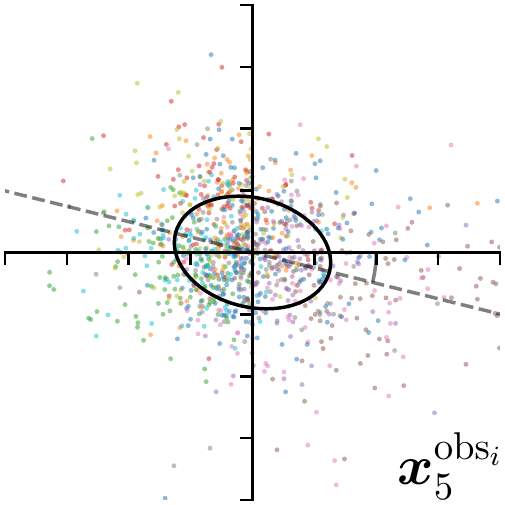} & 
    \includegraphics[width=\linewidth]{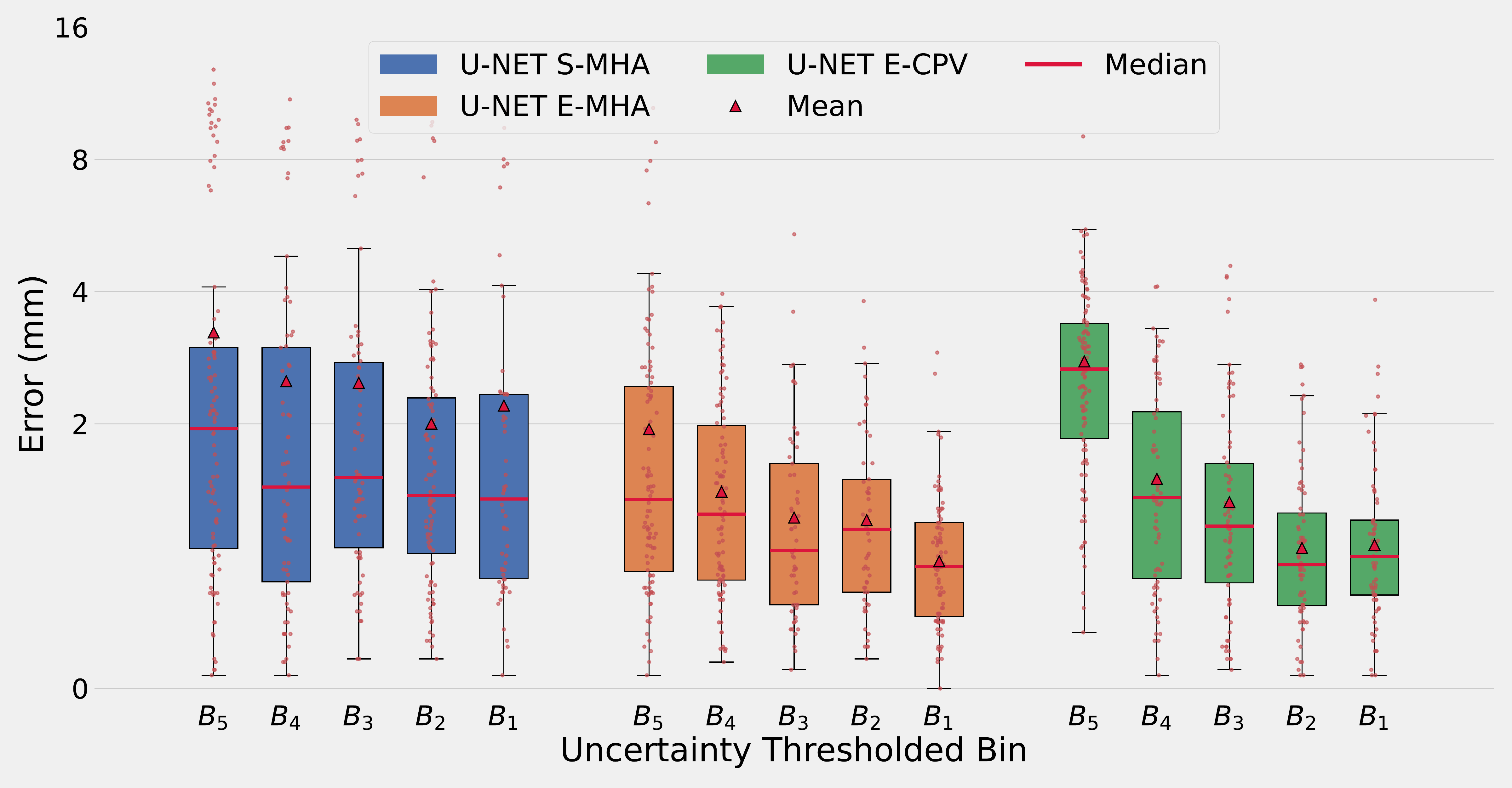}  & 
    \includegraphics[width=\linewidth]{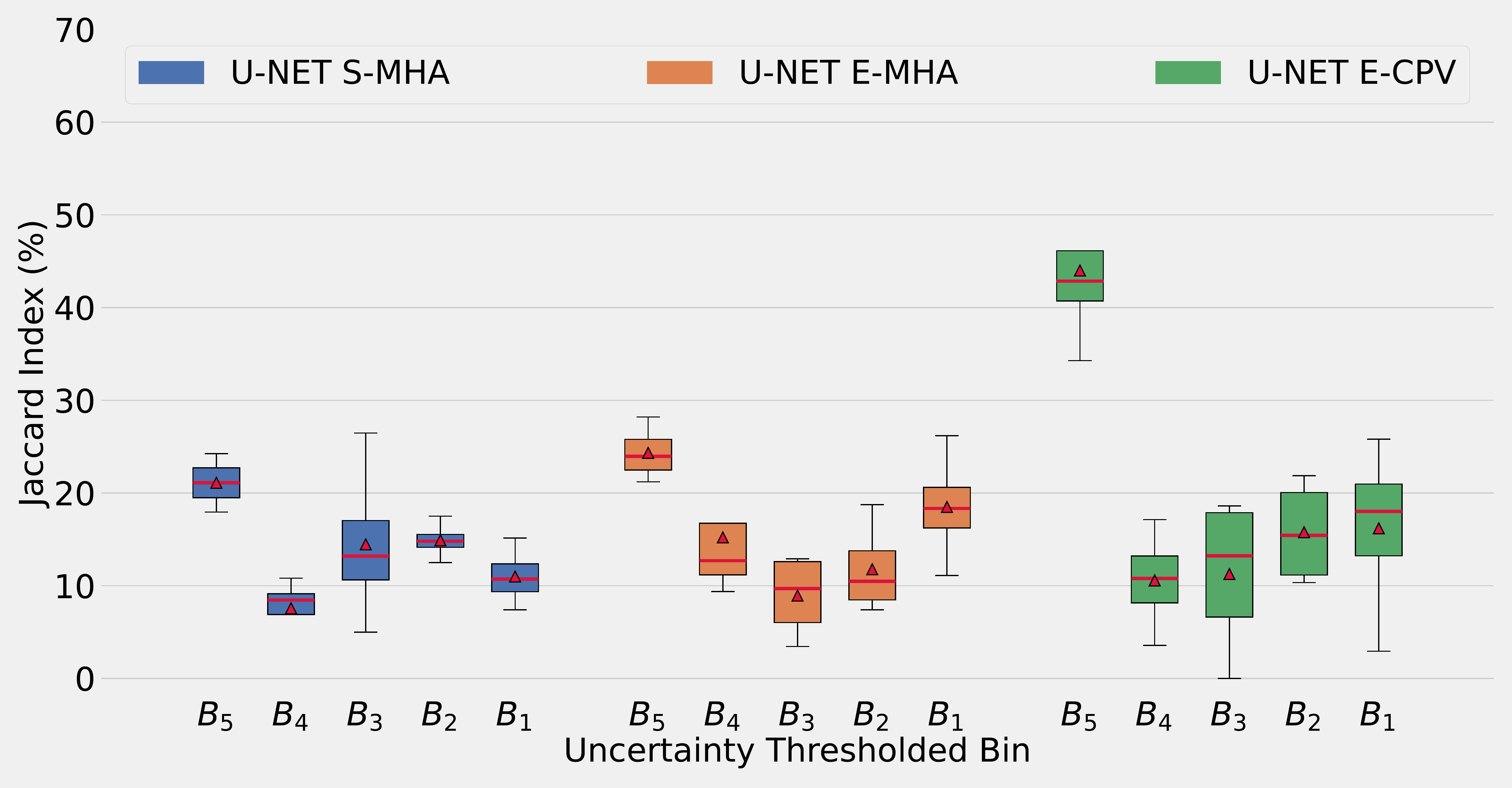} \\
    
    \rotatebox[origin=c]{90}{$L_{3}$} &
    \includegraphics[width=\linewidth]{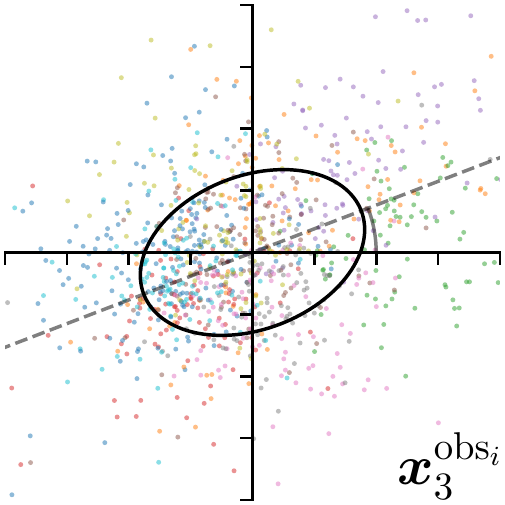} & 
    \includegraphics[width=\linewidth]{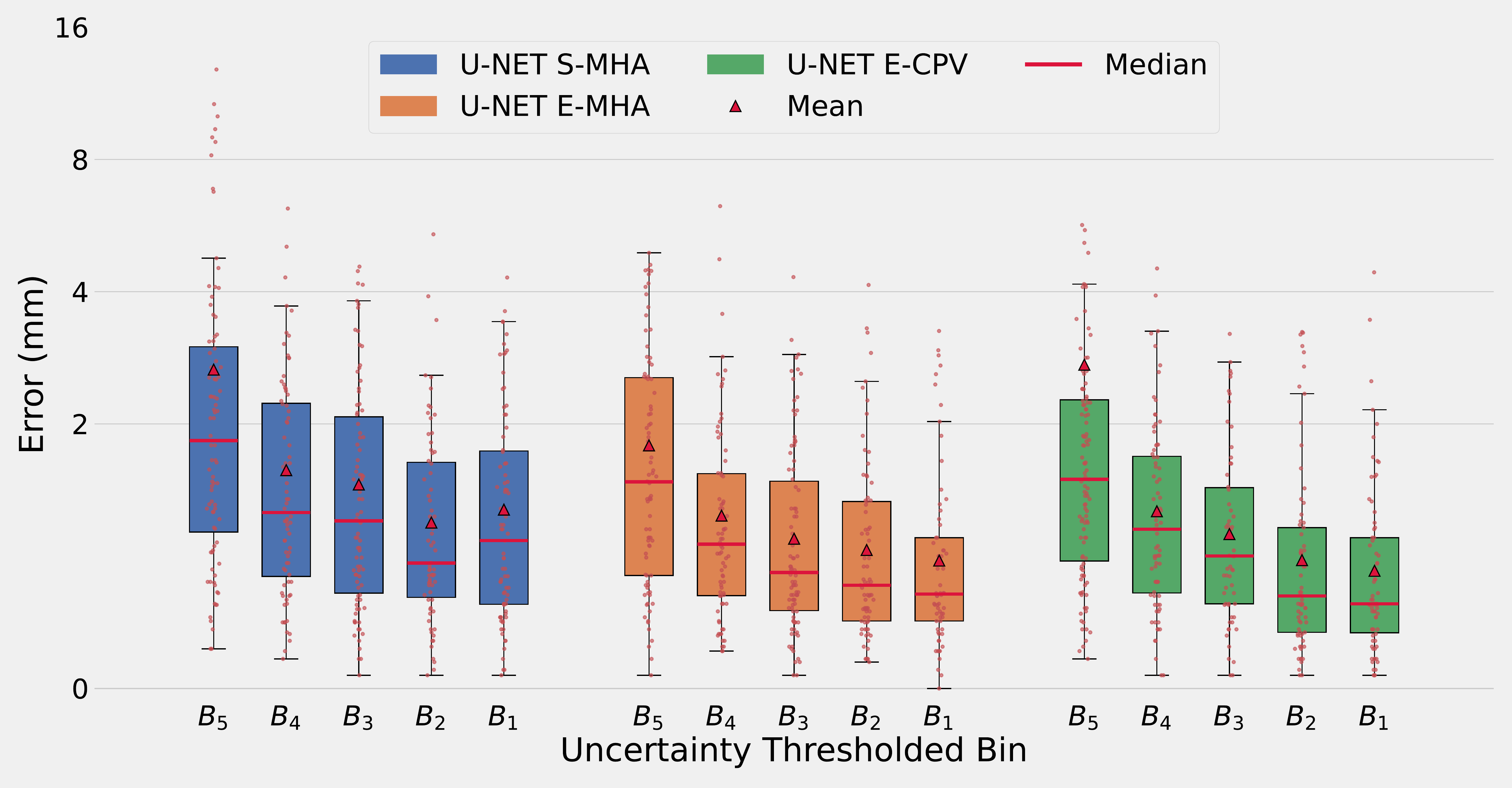}  & 
    \includegraphics[width=\linewidth]{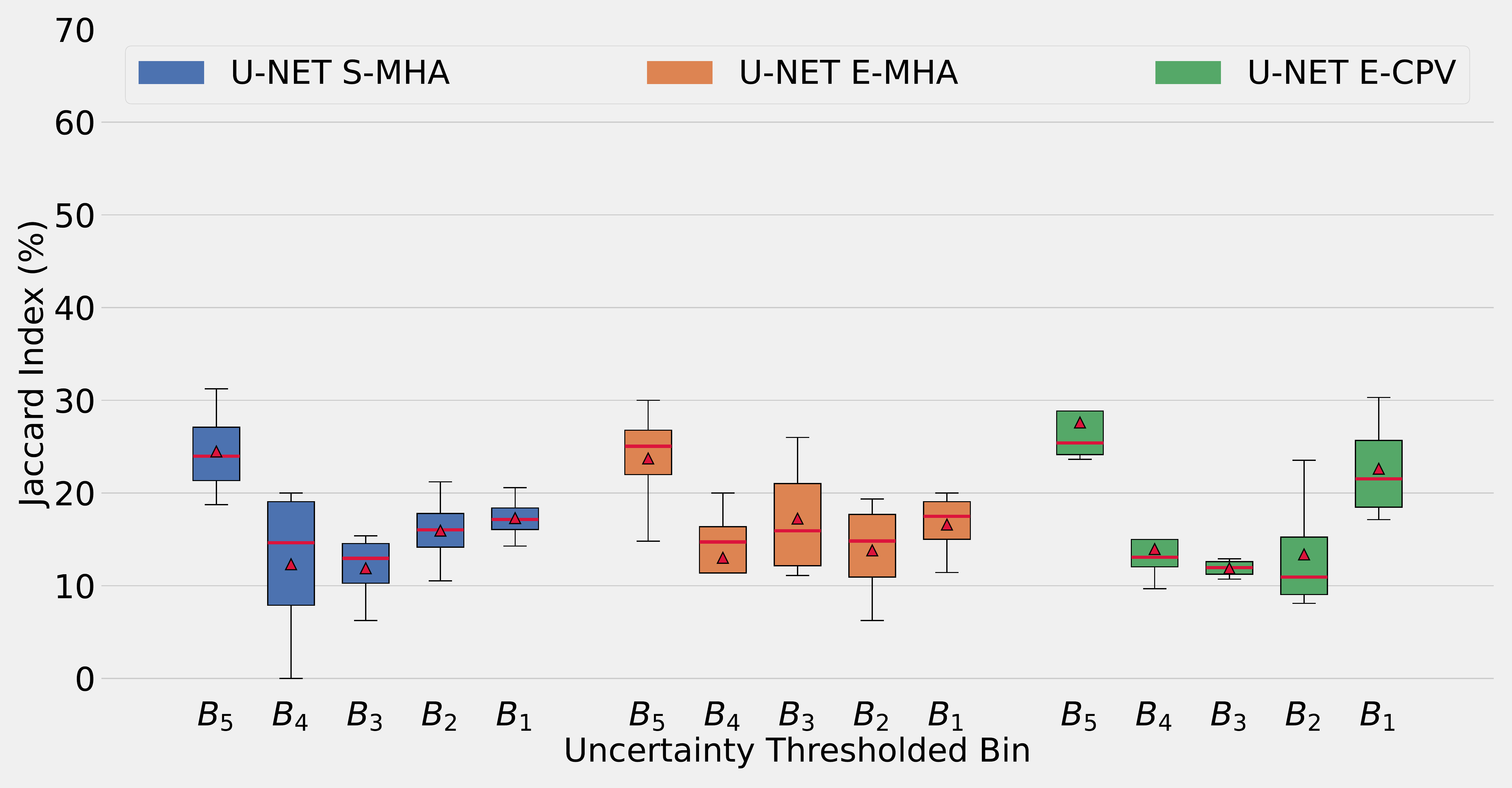}    \\
    
    \rotatebox[origin=c]{90}{$L_{2}$} & 
    \includegraphics[width=\linewidth]{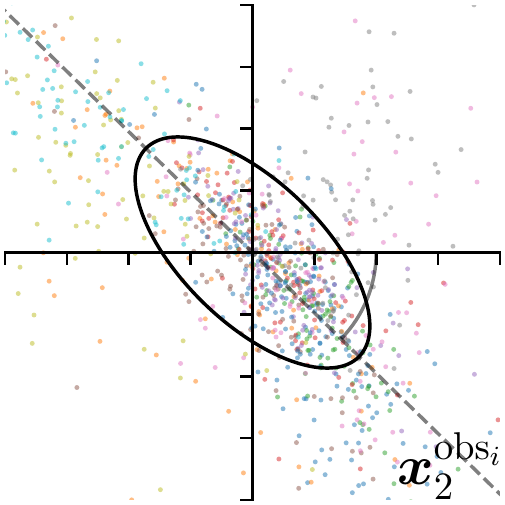} & 
    \includegraphics[width=\linewidth]{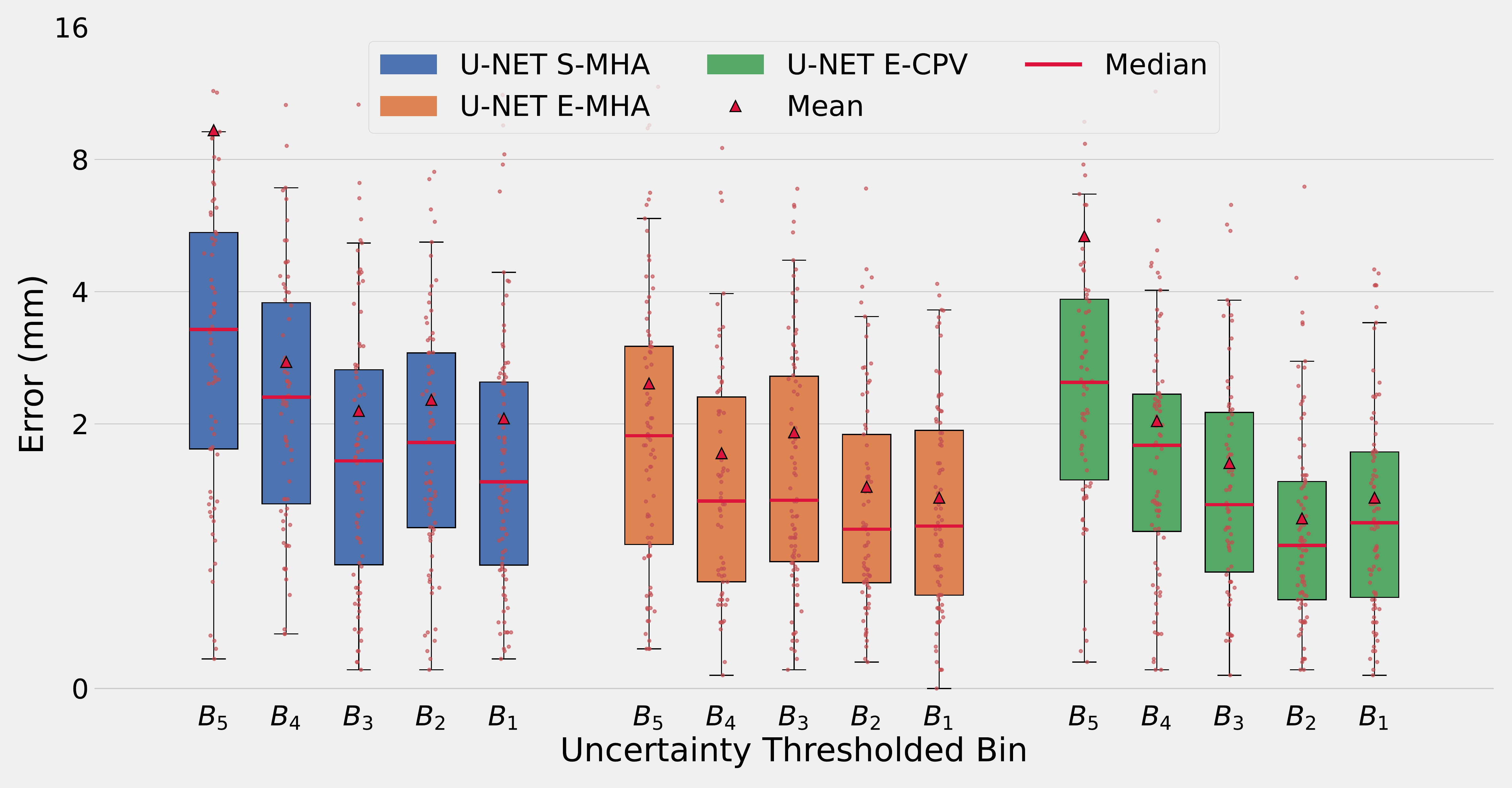}  & 
    \includegraphics[width=\linewidth]{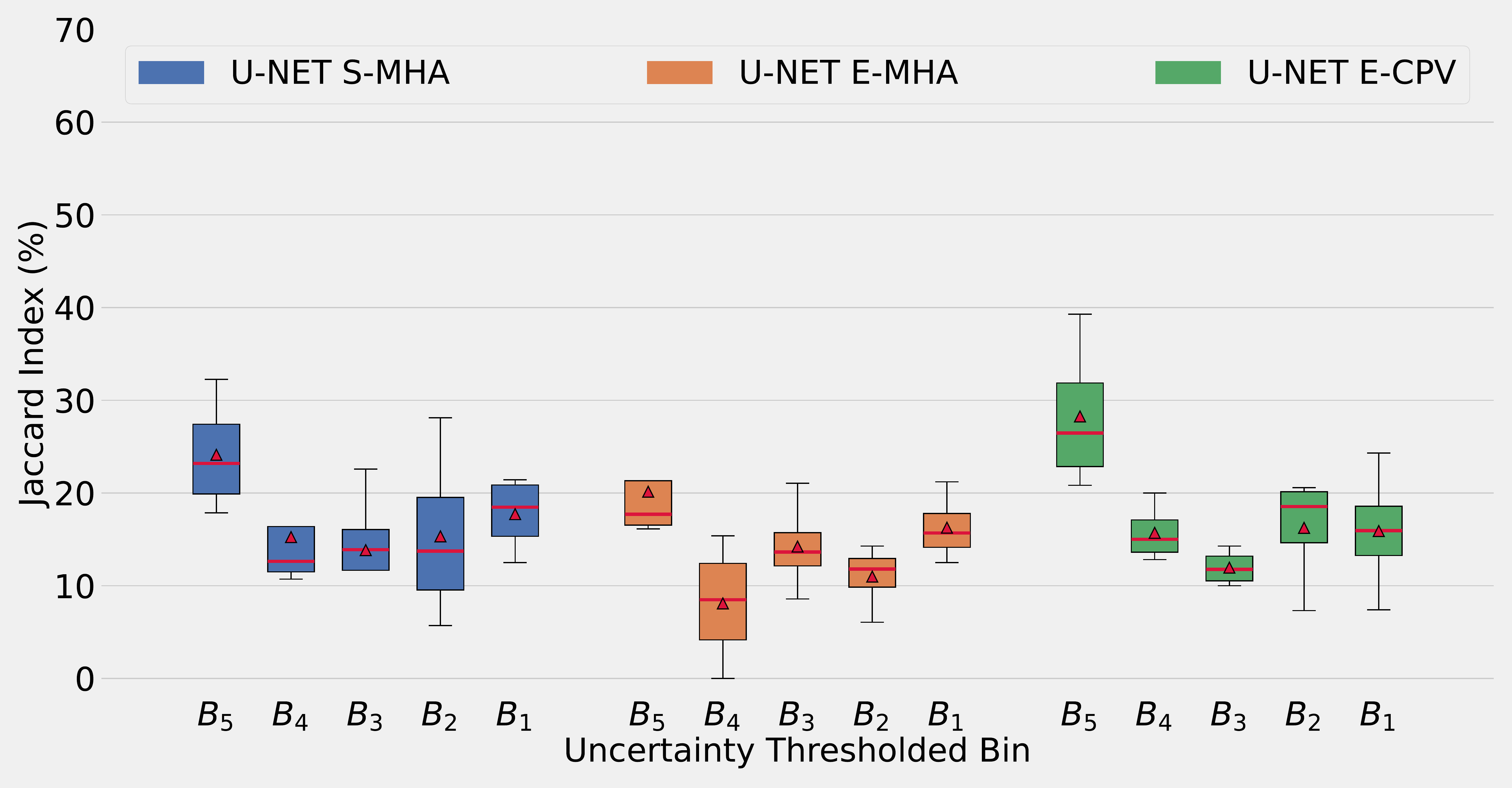}   \\

\end{tabular}

\caption{{\color{red} Column \textit{Annotator Dist.} shows the individual offsets from each of the 11 annotators to the mean annotation of each landmark \cite{thaler2021modeling}. The larger the fitted Gaussian, the more variance between annotators and the higher the aleatoric uncertainty. \textit{Quantile Errors} column shows the boxplots of localization errors for each quantile bin, showing the landmarks across all folds. The \textit{Jaccard Index} column shows the similarity between the predicted Quantiles and the true error quantiles.}}
\label{fig:aleatoric_plots_full_all}
\vspace{-6mm}

\end{figure*}

\vspace{-2mm}

{ \color{red}
\subsection{Relationship with Aleatoric Uncertainty}
\vspace{-1mm}
\label{section:aleatoric_uncertainty} 
Lastly, we study aleatoric uncertainty, which refers to uncertainty caused by internal randomness in the data. Using Quantile Binning, we explore how our epistemic uncertainty measures deal with landmarks of varying levels of aleatoric uncertainty. In landmark localization, one way to measure aleatoric uncertainty is from the inherent ambiguity of the landmark, quantified by the inter-observer variability of multiple annotators. We can infer that the higher the variation in annotator opinion, the greater the ambiguity of the landmark. We can observe the directional ambiguity of the landmark by fitting an anistotropic (directionally skewed) Gaussian function to the distribution of the annotations, seen in the \textit{Annotator Dist.} column of Fig. \ref{fig:aleatoric_plots_full_all}. Thaler \textit{et al.} \cite{thaler2021modeling} provide this ground truth measure of the aleatoric uncertainty, using a total of 11 annotators to label a subset of five landmarks (Fig. \ref{subfig:landmarkisbi}) from 100 images of the Cephalometric dataset. We assume the landmark-specific ambiguities hold for the full Cephalometric dataset.

Fig. \ref{fig:aleatoric_plots_full_all} shows that all studied coordinate extraction methods are best for landmarks with low aleatoric uncertainty. The mean errors ($\blacktriangle$) over the boxplots in the \textit{Quantile Errors} column in Fig. \ref{fig:aleatoric_plots_full_all} confirm that landmarks with higher aleatoric uncertainty ($L_{3}$, $L_{2}$) had worse localization performance than landmarks with low aleatoric uncertainty ($L_{4}$, $L_{1}$). However, the distribution of individual samples (represented by red dots, best seen on screen) show that E-MHA and E-CPV reliably capture the majority of gross mispredictions ($B_{5}$) regardless of landmark ambiguity. S-MHA performs poorly on some landmarks ($L_{5}, L_{2}$) due to the reliance on a single model capacity. In terms of filtering out poor predictions, we see the best results for all uncertainty methods for the landmark with the tightest annotation distribution ($L_{4}$), with $B_{5}$ Jaccard Index's showing a mean of 40\% and 45\% similarity with the true quantile bin for E-MHA and E-CPV, respectively.

However, MHA methods falter for landmarks with directional ambiguity, whereas E-CPV estimates uncertainty well for all types of ambiguity. The \textit{Annotator Dist.} column of Fig. \ref{fig:aleatoric_plots_full_all} shows that the annotation distribution of landmarks $L_{1}$ and $L_{2}$ have a distinct directional skew. The Jaccard Indexes of E-MHA for these landmarks ($L_{1}=16\%$, $L_{2}=20\%$ for $B_{5}$) are lower than the other landmarks with more isotropic annotation distributions ($L_{4}=40\%$, $L_{5}=24\%$, $L_{3}=23\%$ for $B_{5}$). Furthermore, the mean and median localization errors do not consistently trend down across bins for E-MHA for the anisotropic landmarks ($L_{1}$, $L_{2}$).

On the other hand, Quantile Binning shows E-CPV is consistently effective regardless of directional ambiguity, with mean Jaccard Indexes for $B_{5}$ no less than 28\% across all landmarks. This is likely because the objective function (Eq. (\ref{equation:gaussian})) encourages the model to predict isotropic Gaussian Heatmaps, which better match isotropic annotator distributions. When we calculate the mean heatmap to extract the peak pixel using E-MHA, the resulting map will still be constrained to the isotropic properties defined by the objective function. This explains why E-MHA even performs well on the ambiguous yet isotropic landmarks $L_{3}$ and $L_{5}$, but poorly on the directionally ambiguous, anisotropic landmarks $L_{1}$ and $L_{2}$. In contrast, E-CPV calculates the variance between peak pixel activations of a group of individual models, where sampling enough independent predictions of an ensemble can effectively approximate the anisotropic distribution. In practice, if the Quantile Error bins for E-MHA show uniformity as they do in $L_{1}$ and $L_{2}$, this is an indication to the user that the landmark may contain some directional ambiguity.

}
\vspace{-3mm}

\section{Discussion and Conclusion}
\label{sec:discussion}

\subsection{Summary of Findings}
\vspace{-1mm}

This paper presented a general framework to assess any continuous uncertainty measure in landmark localization, demonstrating its use on three uncertainty metrics and two paradigms of landmark localization model. We introduced a new coordinate extraction and uncertainty estimation method, E-MHA, offering the best baseline localization performance and competitive uncertainty estimation.

Our experiments indicate that both heatmap-based uncertainty metrics (S-MHA, E-MHA), as well as the strong baseline of coordinate variance uncertainty metric (E-CPV) are applicable to both U-Net and PHD-Net. Despite the two models' distinctly different approaches to generating heatmaps, using the maximum heatmap activation as an indicator for uncertainty is effective for both models. We showed that all investigated uncertainty metrics were effective at filtering out the gross mispredictions ($B_{Q}$) and identifying the most certain predictions ($B_{1}$), but struggled to capture useful information for the intermediate uncertainty bins ($B_{2}$-$B_{Q-1}$).

Our experiments also showed that E-MHA and S-MHA had a surprisingly similar ability to capture the true error quantiles of the best and worst 20\% of predictions (Figs. \ref{subfigure:jaccard_4ch} \& \ref{subfigure:jaccard_sa}), but E-MHA was more consistent with its performance predicting the error bounds of those bins across models (Figs. \ref{subfig:errorbound4ch} \& \ref{subfig:errorboundsa}). This suggests that the correlation with localization error at the head and tail ends of the heatmap distributions are stable across our ensemble of models, but susceptible to variance when fitting our isotonically regressed line to predict error bounds. On the more challenging 4CH dataset, E-CPV broadly remained the strongest method for filtering out the worst predictions, but this trend did not continue in the easier SA dataset (Fig \ref{figure:all_errors}).

In terms of error bound estimation, we found bins $B_{Q}$ and $B_{1}$ could offer good error bound estimates, but the intermediate bins could not (Figs. \ref{subfig:errorbound4ch} \&  \ref{subfig:errorboundsa}). We found all uncertainty methods performed broadly the same: effective at predicting error bounds for $B_{1}$ and $B_{Q}$, but poor at predicting error bounds for $B_{2}$-$B_{Q-1}$. The one exception was PHD-Net using S-MHA, which could not accurately predict error bounds for $B_{1}$ due to the high variance in pixel activations of highly certain predictions. 

{\color{red}
We demonstrated our Quantile Binning and the three uncertainty metrics are generalizable across imaging modalities by reporting effective results on the Cephalometric dataset in Fig. \ref{fig:comparing_q_emha_ecpv_isbi}. Here, we also showed the flexibility of Quantile Binning by varying the number of bins ($Q$), illustrating the trade-off between true error quantile accuracy and binning resolution as $Q$ increases.

Finally, in Sec. \ref{section:aleatoric_uncertainty} we explored the effect of aleatoric uncertainty on our chosen epistemic uncertainty measures, using Quantile Binning to uncover weaknesses of E-MHA when dealing with landmarks with high directional ambiguity under conventional isotropic heatmap regression.
}
\vspace{-2.7mm}

\subsection{Recommendations}
\label{sec:recommendations}
\vspace{-1mm}
{\color{blue}
We offer the following recommendations:
\begin{itemize}
    \item When resources are available, E-MHA should be used as the coordinate extraction and uncertainty estimation method since it offers the best baseline localization performance with a sufficient ability to filter out the gross mispredictions.
    
    \item If the definition of the landmark is known to be directionally ambiguous, use E-CPV over E-MHA for uncertainty estimation. If this is unknown, uniformity in the E-MHA Quantile Bins can be an indication of directional ambiguity in the landmark. 
    
     \item When resources are constrained, S-MHA is surprisingly effective at capturing the true error quantiles for bins $B_{1}$ and $B_{Q}$, but note that when using a patch-based voting heatmap that is not strictly bounded, the error bound estimation for $B_{1}$ is not robust. 
     
     \item The number of Quantile Bins used ($Q$) is a trade-off, with a larger $Q$ offering a finer binning resolution at the cost of less accurate bins. $Q$ is constrained by the size of the hold-out validation set and can perform poorly when $Q>10$ and the validation set is smaller than 60 samples. 

\end{itemize}
}
\vspace{-4mm}

\subsection{Conclusion}
\vspace{-1mm}

Beyond the above recommendations, we hope the framework described in this paper can be used to assess refined or novel uncertainty metrics for landmark localization, and act as a baseline for future work. Furthermore, we have shown that both the voting derived heatmap of PHD-Net, and the regressed Gaussian heatmap of U-Net can be exploited for uncertainty estimation. In this paper, we only explored the activation of the peak pixel, but it is  likely that more informative measures can be extracted from the broader structure of the heatmap, promising greater potential for uncertainty estimation in landmark localization waiting to be uncovered.

\vspace{-3mm}
\bibliographystyle{IEEEtran}
\bibliography{strings}

\end{document}